\documentclass[journal,twocolumn]{IEEEtran}
\usepackage{xcolor}
\usepackage{mathtools}
\usepackage{epsfig,makeidx,color}
\usepackage{amsmath,amssymb,bbm,enumitem}
\usepackage{cite,graphicx,lipsum}
\usepackage[switch,pagewise]{lineno}
\usepackage{tabularx}
\usepackage{algorithm}
\usepackage{algpseudocode}
\usepackage{hyperref}
\usepackage{cases}
\usepackage[switch,pagewise]{lineno}
\usepackage{graphicx}
\usepackage{booktabs, multirow}
\usepackage{subcaption}
\usepackage{cleveref}
\usepackage{tcolorbox}

\usepackage{subcaption}

\hypersetup{
        colorlinks = true,
        citecolor = red,
        breaklinks = true,
        linktocpage=true,
}
\usepackage{url}

\def\cE{{\cal E}}

\def\cH{{\cal H}}
\def\cK{{\cal K}}

\def\cG{{\cal G}}

\def\rT{{\rm T}}

\def\uR{{\mathbb R}}

\def\be{ \begin{equation} }
\def\ee{ \end{equation} }
\def\bea{ \begin{eqnarray} }
\def\eea{ \end{eqnarray} }

\def\bx{{\bf x}}
\def\by{{\bf y}}

\def\bg{{\bf g}}

\def\bu{{\bf u}}

\def\bh{{\bf h}}

\def\bA{{\bf A}}

\def\bD{{\bf D}}

\def\bF{{\bf F}}

\def\bJ{{\bf J}}
\def\bK{{\bf K}}
\def\bL{{\bf L}}

\def\bW{{\bf W}}
\def\bU{{\bf U}}
\def\bX{{\bf X}}
\def\bY{{\bf Y}}

\def\b0{{\bf 0}}

\def\bSigma{{\bf \Sigma}}

\def\cA{{\cal A}}

\def\cE{{\cal E}}
\def\cL{{\cal L}}
\def\cD{{\cal D}}

\def\cN{{\cal N}}

\def\cV{{\cal V}}

\ifCLASSOPTIONonecolumn
\else

\fi

\begin{document}

\title{Learning Time-Varying Graph Signals via Koopman}


\author{Sivaram Krishnan, Jinho Choi, and Jihong Park
\thanks{S. Krishnan and J. Choi are  with the School of Electrical and Mechanical Engineering, The University of Adelaide, Australia (Emails:\{jinho.choi,sivaram.krishan\}@adelaide.edu.au), and J. Park is with the Information Systems Technology and Design Pillar, Singapore University of Technology and Design, Singapore (Email: jihong\_park@sutd.edu.sg).}
}

\maketitle

\begin{abstract}
A wide variety of real-world data, such as sea measurements, e.g., temperatures collected by distributed sensors and multiple unmanned aerial vehicles (UAV) trajectories, can be naturally represented as graphs, often exhibiting non-Euclidean structures. These graph representations may evolve over time, forming time-varying graphs. Effectively modeling and analyzing such dynamic graph data is critical for tasks like predicting graph evolution and reconstructing missing graph data. In this paper, we propose a framework based on the Koopman autoencoder (KAE) to handle time-varying graph data. Specifically, we assume the existence of a hidden non-linear dynamical system, where the state vector corresponds to the graph embedding of the time-varying graph signals. To capture the evolving graph structures, the graph data is first converted into a vector time series through graph embedding, representing the structural information in a finite-dimensional latent space. In this latent space, the KAE is applied to learn the underlying non-linear dynamics governing the temporal evolution of graph features, enabling both prediction and reconstruction tasks.
\end{abstract}


\ifCLASSOPTIONonecolumn
\baselineskip 28pt
\fi

\section{Introduction}



\subsection{Motivation}
Graphs are fundamental data structures for modeling the structure and interactions within complex systems \cite{Estrada2016-dp} across a variety of domains, including, but not limited to, social networks \cite{Henry06}, biological systems \cite{Li2022}, transportation networks \cite{Huang16}, and communication systems \cite{Suarez23}. These data structures provide a versatile framework for representing relationships and dependencies, enabling insights into the organization and behavior of complex systems. In many real-world applications, the underlying graph data is not static; instead they evolve over time. \emph{Time-varying graphs} \cite{Nicosia12} are a type of graph data characterized by temporal variations in their components or overall configuration. Unlike the commonly studied static graph structures, analyzing time-varying graph data introduces additional challenges. In this work, we consider two application scenarios involving time-varying graphs, as follows: 
\begin{itemize}
    \item the reconstruction of graph signals
    \item the prediction of graph signals.
\end{itemize}  
While the reconstruction of graph signals is necessary for recovering missing information, which is common in real-world sensor networks or data transmission scenarios, prediction, on the other hand, enables forecasting the future states of the systems and thus supports planning, decision-making, and control in dynamical environments.

While several methods exist in the literature for graph signal reconstruction, which is based on the graph signal processing (GSP) \cite{Qiu17, giraldo2022reconstruction} or graph neural network (GNN) paradigm \cite{castro2024gegenbauer}, most of the formulation is limited to time-varying graphs with a fixed structure. While another for graph signal prediction is that most of the data-driven methods are computationally heavy and often fail to capture the non-linear dynamics inherent in time-varying graphs. 

To address these limitations, we propose a novel framework leveraging the \emph{Koopman operator theory} \cite{Koopman31}. This theory demonstrates that non-linear dynamics can be represented linearly by focusing on functions (observables) of the system's state, enabling linear modeling in a higher-dimensional latent space. The Koopman autoencoder (KAE) \cite{Koopman31, Brunton16, lusch2018deep} offers a data-driven approach to linearize complex non-linear dynamical systems by learning such linear representations in a lifted (latent) space. This physics-informed approach ensures predictions adhere to a consistent and interpretable dynamic structure, rather than merely fitting data in a black-box manner. Finite-dimensional approximations \cite{Mezic2005-kd} and Koopman invariant subspaces \cite{Brunton_PLOS} have further advanced this field, with comprehensive reviews available in \cite{Mezi2021, Brunton22} and deep learning applications explored in \cite{Takeishi17, lusch2018deep}.

Building upon our previous work on Koopman-theoretic graph embeddings and the graph KAE (GKAE) \cite{krishnan2025predictive, KrishnanICASSP25}\footnote{This paper is an extended version of \cite{KrishnanICASSP25}, providing a generalized framework for prediction and reconstruction with comprehensive exposition and extensive experimental validation.}, we introduce a robust framework for predicting time-varying graphs and reconstructing missing graph data. Our approach assumes the observed graph sequence arises from an underlying, unknown non-linear dynamical system. For learning on these graphs, the data is first embedded into a latent space (graph embedding) as a multivariate time series, capturing structural information. The GKAE then learns the linear dynamics governing the evolution of these graph embeddings, enabling both tasks. 

The contributions of this paper are summarized as follows:
\begin{itemize}
\item[C1)] We formally categorize time-varying graphs into two broad types: those with time-varying graph signals and those with time-varying graph structures. We highlight the limitations of existing GSP-based and GNN-based approaches for reconstructing time-varying graph signals, which is majorly limited to only time-varying graph signals with a static graph structure.

\item[C2)] We propose the GKAE as a robust framework for reconstruction and prediction in time-varying graphs, capable of handling diverse types of time-varying graphs. The GKAE leverages GNN and Koopman theory, operating in the latent space to model non-linear dynamics through a linear Koopman matrix, making it applicable to all types of time-varying graphs.  

\item[C3)] We evaluate our method against established baselines. For reconstruction tasks, we compare it with GSP among other approaches, while for prediction tasks, we benchmark it against other data-driven Koopman-based methods and GNN-based state-of-the-art methods. Our results demonstrate superior performance in terms of both reconstruction error and prediction error using GKAE across various types of time-varying graphs.
\end{itemize}
\begin{table}[h!]
    \centering
    \caption{Glossary of Notations}
    \begin{tabular}{|c|p{5cm}|}
        \hline
        \textbf{Symbol} & \textbf{Description} \\ 
        \hline
        \hline
        \multicolumn{2}{|c|}{\textbf{Graph Notations}} \\
        \hline
        \hline
        $G(t)$ & Graph at time step $t$. \\
        $\cV$ & Set of nodes. \\
        $\bx(t)$ & Graph signals at time step $t$. \\
        $\cE(t)$ & Edge set at time step $t$. \\
        $\bA(t), \bW(t)$ & Binary/Weighted Adjacency matrix at time $t$.\\
        $\cN_l(t)$ & Neighborhood of node $l$ at time $t$. \\
        $\bL(t) = \bD(t)-\bA(t)$ & Laplacian matrix \\
        \hline
        \hline
        \multicolumn{2}{|c|}{\textbf{Model Notations}} \\
        \hline
        \hline
        $\bg(t)$ & Graph embedding/state vector at time $t$. \\
        $\bK$ & Koopman matrix of dimension $M$. \\
        $\bh(t)$ & Koopman variable at time $t$. \\
        $L$ & Maximum prediction step during training. \\
        $\cE, \cD$ & Encoder and decoder function. \\
        $\theta_{\text{GNN}}, \theta_{\text{KAE}}, \theta_{\text{LC}}$ & Parameters for encoder functions. \\
        $\phi_{\text{GNN}}, \phi_{\text{KAE}}, \phi_{\text{LC}}$ & Parameters for decoder functions. \\
        \text{POOL}$(\cdot)$ & Pooling function. \\
        $\bY$ & Sampled graph signals. \\
        $\bJ$ & Sampling matrix. \\
        $\by'(t)$ & Sampled graph signals at time $t$. \\
        $\bW'(t)$ & Sampled adjacency matrix at time $t$. \\
        \hline
        \hline
        \multicolumn{2}{|c|}{\textbf{Loss Functions and Performance Metrics}} \\
        \hline
        \hline
        $\cL_{\text{GKAE}}$ & Loss function for the graph Koopman autoencoder. \\
        $\cL_{\text{LC}}$ & Loss function for the Latent consistency autoencoder. \\
        $\epsilon_{\text{pred}}$ & Average prediction error. \\
        $\epsilon_{\text{reco}}$ & Average reconstruction error. \\
        $P$ & Total number of prediction steps after training. \\
        $\beta_1, \beta_2$ & Loss hyperparameters. \\
        \hline
        \hline
        \multicolumn{2}{|c|}{\textbf{Dataset}} \\
        \hline
        \hline
        $N$ & Total number of nodes. \\
        $k$ & Number of neighbors using $k$-nearest neighbors. \\
        $r$ & Communication radius for determining the neighborhood. \\
        \hline
        \hline
    \end{tabular}
    \label{tab:glossary}
\end{table}
The comprehensive glossary of notations used throughout the paper is summarized in Table.~\ref{tab:glossary}. 

\subsection{Related Work}
\label{section:related}
We begin by reviewing methods developed for graph signal reconstruction, with a particular emphasis on approaches grounded in GSP. We then examine prediction methods—often used interchangeably with forecasting in the literature—which are majorly data-driven approaches, particularly deep learning and GNN-based models
\paragraph{Reconstruction of Graph Signals}
GSP extends traditional signal processing concepts to graph-based data \cite{Shuman13} \cite{Sandryhaila13}.
Unlike traditional signal processing, which operates on regular grids with established frameworks like Fourier transforms and sampling theories (e.g., the Nyquist rate), GSP addresses signals defined on non-uniform and often dynamic graph structures that represent underlying networks \cite{ortega2018graph} \cite{Leus23}. GSP relies on spectral analysis through operations such as the graph Fourier transform, focusing on processing graph signals in their native domain. However, GSP faces computational challenges, particularly when dealing with large or time-varying graphs. 

GSP-based methods have traditionally been developed with a focus on the graph signal reconstruction problem. For instance, the problem of time-varying graph signal reconstruction is addressed in \cite{Qiu17, giraldo2022reconstruction}. In \cite{Qiu17}, the definition of smooth signals was extended from static to time-varying graph signals, and an optimization problem was formulated for their reconstruction. In contrast, \cite{giraldo2022reconstruction} generalized this formulation by incorporating Sobolev smoothness. The formulation of these methods are limited to time-varying graphs with a static structure. To address this, a notable robust method was proposed in \cite{yamagata2022robust} for recovering time-varying graph signals from corrupted data in dynamic physical sensor networks, where it is assumed that the time-varying graph structure is known a priori. The core contribution is a constrained convex optimization framework designed to handle missing values and outliers. This is achieved by incorporating both graph Laplacian regularization, which enforces signal smoothness across the known dynamic graph, and temporal regularization, which promotes smoothness over time. The problem is then solved efficiently using a primal-dual splitting method. However, this method relies on the assumption of signal smoothness, similar to other approaches in \cite{Qiu17, giraldo2022reconstruction}, which may not be valid for data obtained from complex real-world systems. Additionally, the temporal regularization techniques typically require complete knowledge of the signal, and the associated graph Laplacian is assumed to be known over time. A broader challenge lies in analyzing time-varying graph data directly in the graph domain, as this involves dealing with the high dimensionality of non-Euclidean spaces.  

On the other hand, \emph{graph embedding} methods \cite{GOYAL18} \cite{Xu21} offers a  fundamentally different approach. Graph embedding methods first map the non-Euclidean graph data into a lower-dimensional latent space while preserving its structural properties. While GSP relies heavily on spectral analysis of the graph and operates directly in the node or edge domain, graph embedding methods aim to learn compact vector representations of nodes, edges, or the entire graph. These representations capture both local and global graph information, allowing for more efficient processing of complex, large-scale, or time-varying graphs. Several graph embedding methods exist that vary over how they handle the input, the type of output, and how the information propagates through the graph. Spectral embedding method \cite{belkin2003laplacian} leverages the eigenvalues and the eigenvectors of graph-related matrices for determining a lower-dimensional Euclidean space, capturing the global structure of the graph and local neighborhoods of the nodes. Other methods \cite{perozzi2014deepwalk, tang2015line, grover2016node2vec} use random walks for capturing the neighborhood of the nodes in a graph. While matrix factorization methods \cite{ou2016asymmetric} focus on preserving the high-order proximity by factorizing a general similarity matrix. With the success of neural networks, GNN-based methods are the state-of-the-art for processing graph-structured data \cite{Dai16}. Prominent examples include the seminal works on graph convolutional networks (GCNs) \cite{kipf2016semi} and graph attention networks (GATs) \cite{velivckovic2017graph}, which leverage convolution and attention mechanisms, respectively, to learn powerful representations of graph data.  Although GCN and GAT are effective for learning feature representations on graphs, they often perform poorly when applied to graph signal reconstruction tasks as a standalone architecture. In \cite{chen2021time}, graph unrolling networks are employed for graph signal reconstruction. A known characteristic of such iterative methods is that they often require a high number of parameters to achieve convergence. While in \cite{castro2023time, castro2024gegenbauer}, the authors propose the Gegenbauer-GNN (GegenGNN), which is an encoder-decoder architecture using cascaded Gegenbauer polynomials filters in its convolution layer. The loss function for training includes the Sobolev smoothness regularization, as in \cite{giraldo2022reconstruction}. It is important to note that these aforementioned GNN-based methods were originally formulated for time-varying graphs with a fixed structure. 

Beyond reconstruction, the effective forecasting of time-varying graph signals presents a non-trivial task, which we now review.

\paragraph{Forecasting of Graph Signals} As our method builds upon Koopman theory, we firstly review existing Koopman-based techniques. Although several Koopman-based methods exist \cite{lusch2018deep, azencot2020forecasting, Choi24}, none have been adapted to handle time-varying graphs. In such instances, we have to predict sequentially for each node, which significantly increases the computational complexity or neglect the spatial interactions all together, which leads to performance loss. On the other hand, recurrent neural networks (RNNs) are prominent state-of-the-art architectures in learning on temporal data but they are often computationally heavy and fail to capture long-term dependencies \cite{pascanu2013difficulty}. Other approaches such as long-short term memory (LSTM) \cite{schmidhuber1997long} and gated-recurrent units (GRU) \cite{dey2017gate} have been proposed for ameliorating the aforementioned challenges, inherent in RNN-based architectures. 

Other prominent methods for graph signal forecasting relies on data-driven GNN-based approaches. For instance, spatio-temporal GCN (STGCN) \cite{yu2017spatio} introduces a unified framework that combines spatial graph convolutions with temporal convolutional layers to model both spatial dependencies among nodes and temporal patterns in sequential data. Diffusion convolution RNN (DCRNN) \cite{li2017diffusion} embeds diffusion convolutions within a GRU cell. This enables the model to capture spatio-temporal dependencies by treating the process as a diffusion on the graph that evolves through the recurrent state updates. Other sequence-to-sequence methods employ an encoder-decoder framework that leverages recurrent models to capture temporal dependencies within graph-embedded data. Continuous Product GCN (CITRUS) \cite{einizade2024continuous} performs joint multidomain learning by defining convolutional operations on a Cartesian product graph. This product graph is formed by combining the spatial graph—representing connections among spatial nodes—with a temporal graph that encodes relationships across time steps. The resulting unified graph’s nodes correspond to specific spatial locations at particular time instances. CITRUS applies learnable continuous heat kernels on this product graph, which act as filters modeling how information diffuses smoothly across both spatial and temporal dimensions. 

\paragraph{Main Differences} Unlike conventional overparameterized models, we propose an efficient approach tailored for natural systems. By enforcing a linear latent space, the model learns the system’s underlying dynamics—approximating the Koopman operator—rather than fitting input-output mappings. This yields interpretable predictions grounded in physical structure, analyzable via the spectral decomposition of the learned Koopman matrix.

The structure of the paper is as follows: Section~\ref{sec:TVG} introduces our framework for time-varying graphs and their associated graph signals, categorizing them based on the components of the graph that change over time. It also outlines the limitations of existing approaches. Section~\ref{sec:GKAE} presents our proposed method for both prediction and reconstruction tasks on time-varying graphs. The problem formulations for these tasks are provided in Sections~\ref{sec:preds} and~\ref{sec:recon}, respectively. Section~\ref{sec:SSR} describes the experimental setup, reports the results of our approach, and compares its performance against established baselines. Finally, Section~\ref{sec:C} concludes the paper.

\section{Background: Time-Varying Graphs} \label{sec:TVG}
In this section, we briefly present an overview of time-varying graphs and associated graph signals.
\subsection{Properties of Graphs}
A graph is represented as $G = (\cV, \cE)$, where $\cV = \{1,2,\ldots,N\}$ is  the set of the nodes (over vertices) and $\cE = \{(l,m): \ l,m \in \cV\}$ denotes the set of the edges \cite{Chung97}. Here, $N$ is the number of nodes. The set of neighbors of node $l$ is denoted by $\cN_l = \{m \ | \ (l,m) \in \cE\}$, and for unweighted graphs, the degree of node $l$ is given by $d_l = |\cN_l|$, i.e., the number of neighbors.

For a given graph, the $N \times N$ Laplacian matrix is defined as
\be
\bL = \bD  - \bA,
	\label{EQ5:Lap}
\ee
where $\bD = \mbox{diag}(d_1, d_2, \ldots, d_N)$ and $\bA$ is the $N \times N$  adjacency matrix, which is given by
\be
[\bA]_{l,m} = \left\{
\begin{array}{ll}
1, & \mbox{if $(l,m) \ \mbox{or} \ (m,l) \in \cE$;} \cr
0, & \mbox{otherwise.} \cr
\end{array}
\right.
\ee
For an undirected graph, $\bA$ is symmetric, while $\bA$ is asymmetric for a directed graph. If each node has the same degree, i.e., $d_1 = \ldots = d_N = d$, the graph is called a $d$-regular graph. 

A weighted graph is a graph with edge weights. For an edge in $\cE$, a non-negative weight can be given, which is denoted by $w_{l,m}$. Thus, the weight matrix, denoted by $\bW$, where $[\bW]_{l,n} = w_{l,n}$, can be seen as a weighted adjacency matrix, and a weighted graph is represented as $G = (\cV, \cE, \bW)$.
The degree of a node in a weighted graph is now defined as the sum of the weights of all the edges connected to it. Consequently, the Laplacian matrix of a weighted graph becomes
\be 
\bL = \bD - \bW.
\ee 
If $w_{l,m} = 1$ for $(l,m) \in \cE$, then $\bW = \bA$. Thus, the weighted graph is a generalized version of the conventional graph. For undirected graphs, $\bL$ is symmetric. In addition, the eigenvalues of $\bL$ are nonnegative, meaning that $\bL$ is a positive semidefinite matrix \cite{Chung97}. Thus, the eigendecomposition of $\bL$ becomes 
\be 
\bL = \bU \bSigma \bU^\rT,
\ee 
where $\bSigma$ is the diagonal matrix of the eigenvalues, $\{\lambda_l \ge 0\}$, i.e., $\bSigma = {\rm diag}(\lambda_1, \ldots, \lambda_N)$, and $\bU$ is the matrix whose column vectors are the eigenvectors, i.e., $\bU = [\bu_1 \ \ldots \ \bu_N]$. Here, $\bu_l$ stands for the eigenvector corresponding to $\lambda_l$.


Let $\lambda_n$ denote the $n$th smallest eigenvalue. For a connected graph, we have $\lambda_1 = 0$ and $\lambda_2 > 0$ \cite{Chung97} \cite{Shuman13}, i.e.,
$$
0 = \lambda_1 < \lambda_2 \le \ \cdots \ \le \lambda_N .
$$
Let the volume of the graph be ${\rm vol}(G) = {\rm tr}(\bL) = \sum_{n=1}^N \lambda_n$. Then, it can be shown that
\be 
\lambda_2 \le \frac{{\rm vol}(G)}{N-1},
\ee 
where the equality can be achieved when $\lambda_2 = \ldots 
= \lambda_N$.

\subsection{Graph Signals}
In the context of GSP, a graph signal signifies scalar values over the nodes of a graph such that, $\bx = [x_1 \ \ldots \ x_N]^\rT \in \uR^N$. In \cite{HAMMOND11}, the graph Fourier transform of $\bx$ is defined as
\be 
\hat \bx = \bU^\rT \bx,
\ee 
and since $\bU$ is unitary (for undirected graphs), the inverse graph Fourier transform is given by
\be 
\bx = \bU \hat \bx .
\ee 

In \cite{Shuman13}, 
the following graph Laplacian quadratic form \cite{Spielman19} has been considered for the global smoothness:
\begin{align} 
S_2 (\bx) & = \bx^\rT \bL \bx \cr 
& = \sum_{(l,n) \in \cE} w_{l,n} (x_l - x_n)^2 .
    \label{EQ:S2}
\end{align}
It can be shown that
\begin{align}
S_2 (\bx) & = \bx^\rT \bU \bSigma \bU^\rT \bx \cr 
 & = \hat \bx^\rT \bSigma \hat \bx = \sum_{n=1}^N \lambda_n |\hat x_n|^2,
\end{align}
For a connected graph, since $\lambda_1 = 0$, the graph Laplacian quadratic form is bounded as follows:
\be 
0 \;\le\; S_2(\mathbf{x}) \;\le\; \lambda_N \lVert \mathbf{x} \rVert^2 .
\ee
Moreover, if $\mathbf{x}$ is orthogonal to the constant vector 
(i.e., $\sum_{i=1}^N x_i = 0$), then the tighter bound holds:
\be
\lambda_2 \lVert \mathbf{x} \rVert^2 \;\le\; S_2(\mathbf{x}) \;\le\; \lambda_N \lVert \mathbf{x} \rVert^2 .
\ee


\subsection{Time-Varying Graphs}
\begin{figure}[thb]
\centering
\includegraphics[width=1\columnwidth]{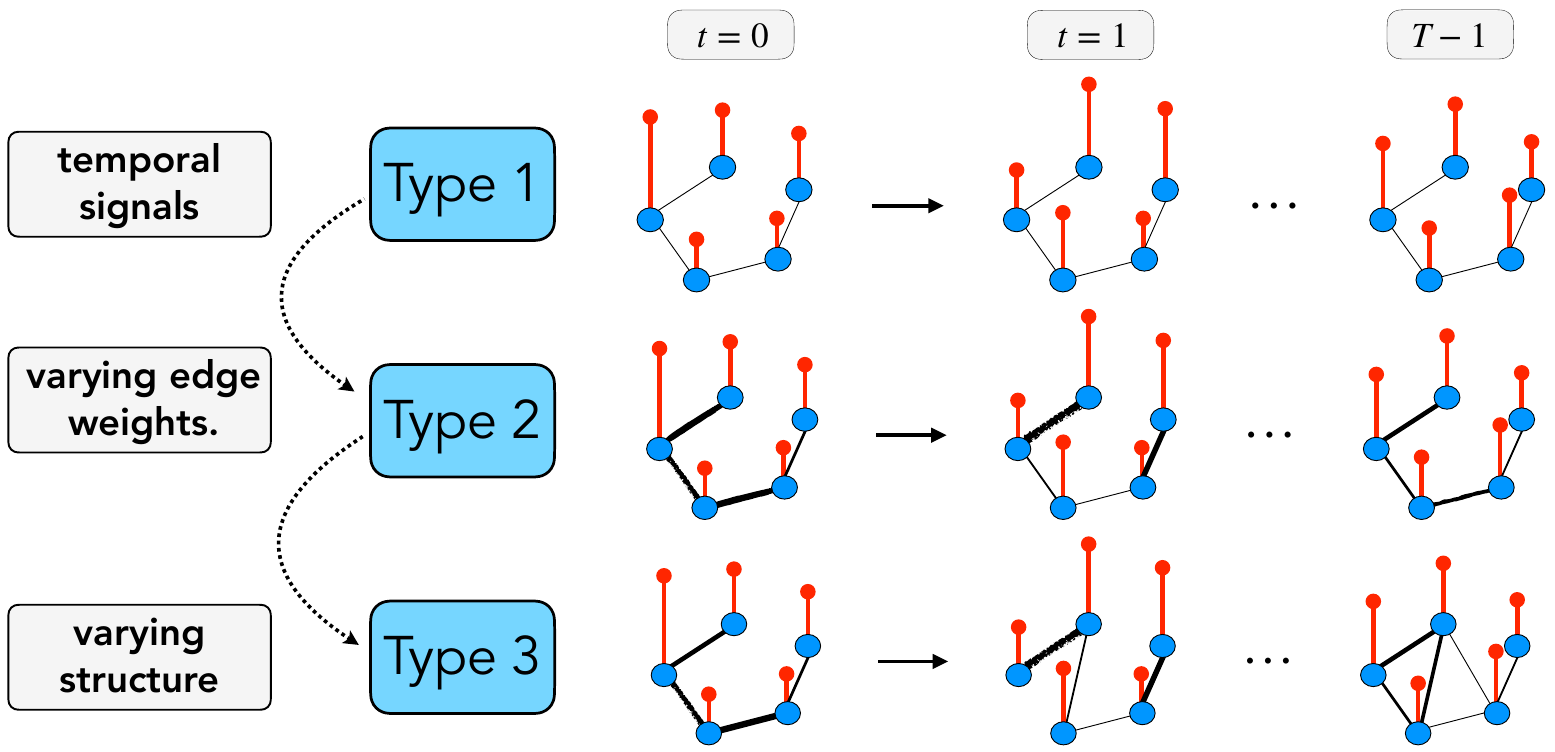}
\caption{Types of time-varying graphs: Type-1 varies in signals only, Type-2 in signals and edge weights, and Type-3 in signals, edge weights, and structure.}
    \label{Fig:Types}
\end{figure}

There are different types of time-varying signals associated with given graphs. Assuming that the number of nodes is fixed (or $\cV$ is fixed), we categorize time-varying graphs into the following three types:
\begin{itemize}
\item {\bf Type-1}: The signals $\bx$ can be extended over time as time-varying graph signals or as a time series, which are referred to as time-varying graph signals \cite{Qiu17}. Throughout the paper, the time-varying graph signal is referred to as type-1. It is noteworthy that the graph structure is time-invariant, i.e., $\cE$ and $\bW$ remain invariant over time, while the node features, i.e., $\bx(t)$, are time-varying.

\item {\bf Type-2}: While the graph structure is invariant over time for type-1 graphs, the edge weights can vary. In type-2 graphs, we assume that $\cV$ and $\cE$ are invariant, while the weights are time-varying, i.e., $\bW(t)$.

\item {\bf Type-3}: 
The connectivity of a graph can vary over time, i.e., $\cE$ becomes $\cE(t)$. The resulting graph is referred to as a type-3 time-varying graph, and the time-varying signal becomes a three-tuple $(\bx(t), \cE(t), \bW(t))$.
According to \cite{Jin2023-gf}, type-1 graphs are referred to as attributed graphs, while type-3 graphs are referred to as spatial-temporal graphs. 
\end{itemize}
In Fig.~\ref{Fig:Types}, we illustrate the relationship between different types of time-varying graphs, where type-1 graphs are a subset of type-2 graphs, and type-2 graphs are a subset of type-3 graphs. Note that if a graph is fully connected, types 2 and 3 are equivalent.

\subsection{Existing Problem Formulation for Graph Signal Reconstruction}
For a finite-length sequence of a graph signal $\bx(t)$, let $\bX = [\bx(0)\ \ldots \ \bx(T-1)] \in \mathbb{R}^{N \times T}$ represent the complete signal. We assume that the data is fully available for the first $\tau$ time steps, after which only a subset of the graph signal is available at each time $t$. Let $ \cA(t)$ denote the index set of nodes with measurements available at time $t$. The available graph signals are then expressed as
\be 
\mathbf{Y} = \mathbf{J} \odot \mathbf{X} \in \mathbb{R}^{N \times T}, \label{eq:subs}
\ee
where $\odot$ represents the Hadamard product and $\bJ \in \mathbb{R}^{N \times T}$ is the sampling matrix, given as 
\be 
[\bJ]_{n,t} = \left\{ 
\begin{array}{ll}
1, & \mbox{if $t < \tau$} \cr
1, & \mbox{if $n \in  \cA(t)$} \cr
0, & \mbox{otherwise.} \cr 
\end{array} \right.
\ee
Note that $\cA(t)$ may vary with time $t$. 


Recent GSP-based methods \cite{Qiu17, giraldo2022reconstruction, yamagata2022robust}, along with GNN-based approaches such as \cite{castro2024gegenbauer}, extend the notion of signal smoothness to the temporal domain \cite{Leus23}. For instance, spatial smoothness can be characterized using the Laplacian matrix, as illustrated in \eqref{EQ:S2}. For time-varying graph signals, temporal smoothness can also be defined in terms of the Laplacian matrix.
To explain this approach, let $\tilde \bx(t) = \bx(t) - \bx(t-1)$ for $t \in {1, \ldots, T-1}$, which denotes the temporal difference signal. From \eqref{EQ:S2}, a set of temporally smooth graph signals, which are called the $\epsilon$-structured time-varying graph signals, can be defined as
\be 
\Omega_\epsilon = \left\{\bX: \ \sum_{t = 1}^{T-1}S_2 (\tilde \bx(t)) \le (T-1)\epsilon\right\} .
    \label{EQ:Oeps}
\ee 
The following optimization is widely considered for time-varying graph reconstruction:
\be
\min_\bX   S_2 (\tilde \bx(t)) + \|\bJ \odot \bX - \bY\|^2_{\rm F},
\label{eq:or}
\ee
where $\|\cdot\|_{\rm F}$ is the Frobenius norm. 


As detailed in Sec.~\ref{section:related}, the methods in \cite{Qiu17, giraldo2022reconstruction, castro2023time, castro2024gegenbauer} are limited to type-1 graphs. Even approaches like \cite{yamagata2022robust}, which target time-varying graphs, assume full knowledge of the dynamic structure and rely on simplistic temporal differences—making them insufficient for capturing complex, non-linear dynamics. Additionally, as the number of time steps $T$ increases, the term \( (T-1)\epsilon \) grows proportionally, permitting larger deviations from smoothness across the entire time horizon and thus adding complexity to the reconstruction process. 

In the following section, we formally introduce our approach, starting with the prediction task. As reconstruction builds upon the predicted signals, we present the reconstruction framework thereafter.

\section{GKAE for Time-Varying Graph Signals: Architecture and Training} \label{sec:GKAE}
In this section, after introducing the Koopman operator and its autoencoder implementation, we present our proposed approach, GKAE. Our model effectively handles non-Euclidean, time-varying graph data (including Type-3 signals) and offers an interpretable approach for predicting and learning their inherent dynamics.
\subsection{Koopman Operator Theory}
The Koopman operator theory  provides a powerful framework for linearizing the evolution of a non-linear dynamical system \cite{Koopman31}
\cite{Budisic2012-zy} \cite{Mezi2021} \cite{Brunton22}. In this subsection, we present an overview of the  Koopman operator theory.

Denote by $\bg(t)$ the state of a non-linear dynamical system that evolves over time as follows:
\be
\bg(t+1) = \bF(\bg(t)) \in \cG \subseteq \uR^N, \label{EQ2:xFx}
\ee
where $\bF$ represents the flow map, a non-linear and unknown function that governs the dynamics of the system and $\cG$ is the state space, which is a subset of the $N$-dimensional vector space.  
According to the Koopman operator theory, given a measurement function, $h: \mathcal{G} \rightarrow \mathbb{R}$, which is an observable function, there exists a linear operator called the \emph{Koopman operator},
denoted by $\mathcal{K}$, which can be applied to all such observable $g$, to advance them in time, as follows:
\begin{align}
\cK h = h \circ \bF,
    \label{EQ3:KO}
\end{align}
where $\circ$ represents the composition operator, existing on a smooth manifold. Applying \eqref{EQ3:KO} to \eqref{EQ2:xFx}, we have
\begin{align}
h(\bg(t+1)) = h \circ \bF(\bg(t)) = \cK h(\bg(t)),
    \label{EQ:gg}
\end{align}
where $h(\bg(t))$ is an observable measured at time $t$.
This can be extended to the case with multiple observables. Precisely, let $\bh (t) = [h_1 (t) \ \ldots \ h_M (t)]^\rT$, where $h_m (t) = h_m (\bg(t))$. Then, from \eqref{EQ:gg}, we have
\be 
\bh(t+1) = \cK \bh(t). 
    \label{eq:linear}
\ee 
We can readily show that the Koopman operator is linear, since 
$\cK (c_1 \bh_1 (t) + c_2 \bh_2(t)) = c_1 \bh_1 (t+1) + c_2 \bh_2(t+1)$, where $c_1, c_2 \in \uR$.

If $\bh(t) \in \cH$ and $\cK \bh(t) \in \cH$, where $\cH$ is a finite-dimensional space, $\cH$ becomes a Koopman invariant subspace \cite{Brunton_PLOS}. In this case, the Koopman operator in \eqref{eq:linear} becomes a square matrix, denoted by $\bK$.  

\subsection{Koopman Autoencoder}
Although the Koopman operator is useful for linearizing non-linear dynamical systems, which aids in modeling and prediction, finding the Koopman invariant subspace is challenging. Data-driven techniques, such as dynamic mode decomposition (DMD) methods \cite{Tu14} \cite{Kutz16} and deep learning based approaches \cite{Takeishi17} \cite{lusch2018deep}, have been developed to address this challenge. In particular, the approaches in \cite{Takeishi17} \cite{lusch2018deep} are based on the autoencoder architecture, and called
the KAE, which consists of the  three main components: an encoder, a Koopman matrix, and a decoder, where the encoder and decoder are neural networks. It is designed to estimate the Koopman operator and its invariant subspace, capturing the smooth dynamics through a linear operator.

Let $\cE_{\theta_{\text{KAE}}} (\bg)$ be the encoder that maps the state vector $\bg$ to a latent variable, denoted by $\bh \in \uR^M$, where $M$ is the dimension of the latent space. Here, $\theta_{\text{KAE}}$ represents the parameter vector for the encoder of the KAE. Then, we assume that the encoder performs the linearization so that the following relation can hold:
\begin{align} 
\cE_{\theta_{\text{KAE}}} (\bg(t+1)) & = \bh (t+1) \cr 
& = \bK \bh (t) = \bK \cE_{\theta_{\text{KAE}}}  (\bg(t)),
    \label{EQ:enc}
\end{align}
where $\bK$ is the Koopman matrix. By allowing the inverse mapping with the decoder, denoted by $\cD_{\phi{\text{KAE}}}$, we can have the original state vector as follows:
\be 
\bg(t) = \cD_{\phi_{\text{KAE}}} (\bh(t)),
    \label{EQ:dec}
\ee 
where $\phi_{\text{KAE}}$ is the parameter vector of the decoder. Here, both the encoder and decoder are assumed to be neural networks. 
From \eqref{EQ:enc} and \eqref{EQ:dec}, using the linearity, we can derive the following relation:
\begin{align}
    \bg (t+l) = \cD_{\phi_{\text{KAE}}} ( \bK^l \cE_{\theta_{\text{KAE}}}  (\bg(t))), \ l = 0, 1,\ldots.  \label{EQ:pred}
\end{align}
This relation holds for all $l \ge 0$ if an encoder can be designed to map the state vector into the Koopman invariant subspace, and the decoder serves as the inverse function of the encoder. According to Koopman operator theory, the dimension of the latent space, $M$, can be infinite. However, in practice, $M$ is finite, meaning the relation in \eqref{EQ:pred} holds approximately and only for a finite number of distinct values of $l$.

Consequently, with a given set of state vectors, $\{\bg(0), \ldots, \bg(T-1)\}$, it may be possible to determine the parameter vectors, $\theta_{\text{KAE}}$ and $\phi_{\text{KAE}}$, and the Koopman matrix, $\bK$, by minimizing the error or loss defined as:
\be 
{\cL}_{\text{KAE}}(\theta_{\text{}}, \phi_{\text{}}, \bK, L) =\sum_{l=0}^{L-1}
\sum_t ||\bg(t+l) - \cD_{\phi_{\text{KAE}}} ( \bK^l \cE_{\theta_{\text{KAE}}} (\bg(t))) ||^2, \label{eq:KAE}
\ee 
where $L$ is a finite positive integer representing the maximum prediction step. 

If the Koopman encoder successfully maps the state vector into the Koopman invariant subspace, \eqref{EQ:pred} holds for any $l \ge 0$ due to global linearization. However, with the finite dimensionality of the latent space, the encoder can only approximate the Koopman invariant subspace.  Consequently, when $L$ is small, the normalized loss, $\frac{1}{L} \cL_{\rm KAE}$, may be small, but the approximation is not as effective. Increasing $L$ ensures a better linearization, as a larger value allows for a more accurate approximation of the Koopman invariant subspace over a longer time horizon.

As mentioned earlier, the dimension of the latent space, $M$, is ideally infinite according to Koopman operator theory and is typically expected to be larger than $N$, the dimension of the state space, especially when modeling non-linear dynamical systems where a high-dimensional latent representation is necessary to capture the full dynamics. However, in some cases, $M$ can be smaller than $N$, resulting in the KAE having an hourglass shape, where the low-dimensional latent space is designed to extract key features or dominant modes of the system, serving as a compact representation of its underlying dynamics \cite{Brunton22}. This dimensionality reduction is advantageous for applications such as modeling multivariate time series, where a simplified yet informative model is sufficient. 

\subsection{GKAE Application 1 : Prediction on Time-varying Graphs}
\label{sec:preds}
The time-varying graph data is represented as a sequence of spatio-temporal graphs
\be
\mathbb{G} = [G(0), \cdots, G(T-1)],
\ee 
where for each time step $t$, $G(t) = (\bx(t), \mathcal{E}(t), \bW(t))$. 
While the traditional KAE is effective in modeling non-linear dynamics using an autoencoder architecture \cite{Takeishi17} \cite{lusch2018deep}, it is typically applied to data originating from a single node, modeled as a Euclidean sequence of input states. This inherently limits its applicability to non-Euclidean data, such as spatio-temporal graphs, where the dynamics arise from interactions among multiple nodes in a graph structure. To extend the modeling capability of the KAE to non-Euclidean sequences, we propose transforming the non-Euclidean graph data into a Euclidean-compatible input. This is achieved by learning a graph embedding that captures the spatial and temporal relationships between nodes at each time step. The graph embedding serves as a compact state vector for each spatio-temporal graph, enabling the KAE to model the dynamics of the sequence effectively while preserving the underlying graph structure and relationships.

\begin{figure*} [thb]
    \centering
\includegraphics[width=0.9\textwidth]{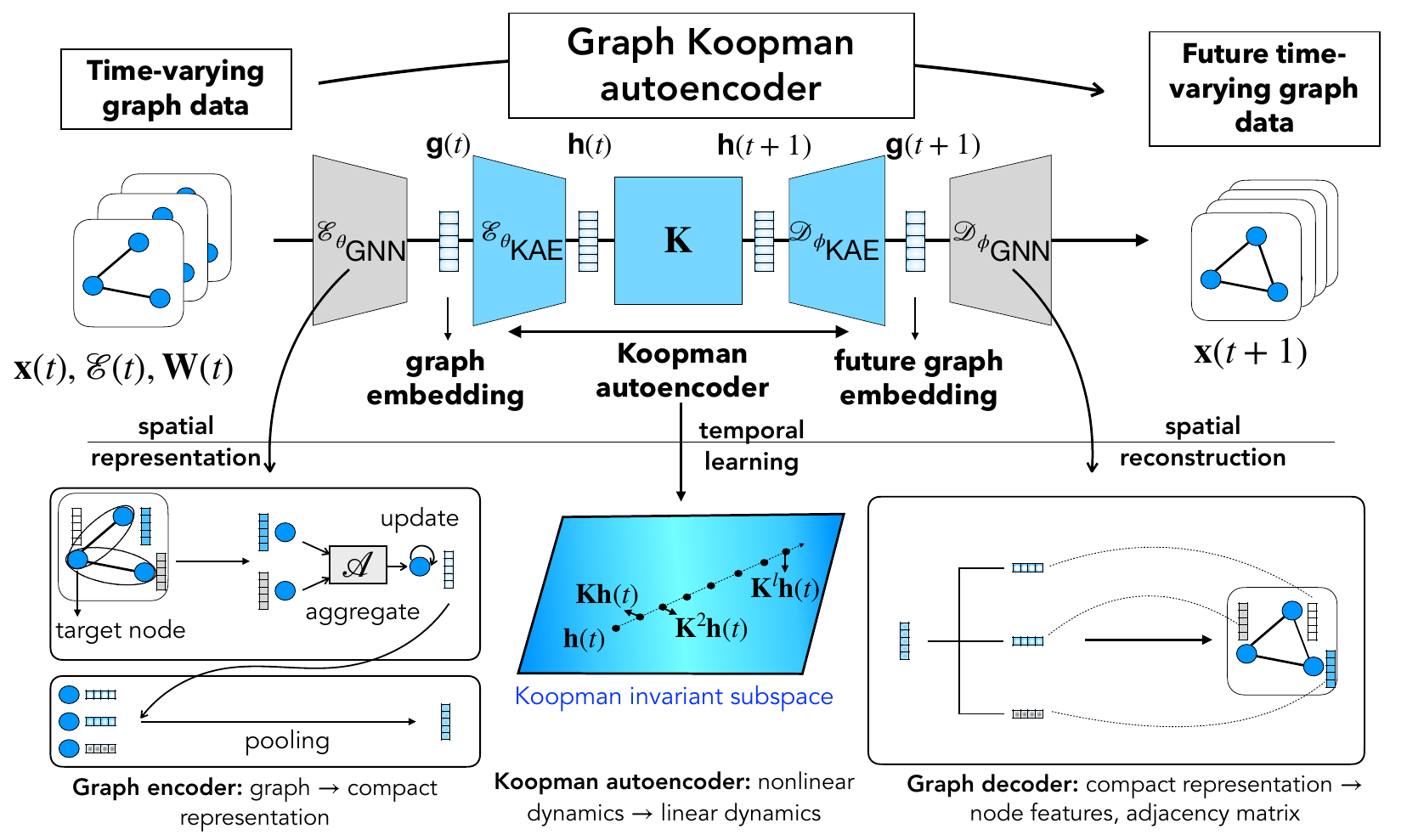}
\caption{Schematic of the proposed Graph Koopman Autoencoder (GKAE) framework. The graph encoder aggregates ($\cA$) and updates graph signals based on neighbors to generate a graph embedding. The Koopman autoencoder predicts future embeddings linearly, and the graph decoder reconstructs the graph in its original space.}    
\label{fig:Proposed}
\end{figure*}

While various techniques exist for converting a graph realization into a graph embedding, we adopt an autoencoder architecture, which has demonstrated significant success \cite{shi2023gigamae, kipf2016variational} in capturing high-dimensional graph structures and representing them as compact graph embeddings. This approach enables the effective transformation of complex graph data into a lower-dimensional latent space while preserving the essential spatial and temporal relationships within the graph. We input the encoder and decoder on either side of the KAE as shown in Fig.~\ref{fig:Proposed}, which is thereafter referred to as the graph encoder and graph decoder, denoted by $\cE_{\theta_{\text{GNN}}}$ and $\cD_{\phi_{\text{GNN}}}$, respectively. The GKAE consists of five components: a graph encoder, a Koopman encoder, a Koopman matrix, a Koopman decoder, and a graph decoder.

The operations within the GKAE can be described using the following components:

\subsubsection{Graph Encoder} The graph encoder decomposes each graph realization as a graph embedding, which aims at finding the latent vectors preserving the spatial characteristics of the graphs. The latent variable allows us to represent the dynamical system of multiple nodes as an Euclidean sequence of input, enabling the use of KAE to linearize the dynamics of the latent variables, which serve as state vectors for the time-varying graphs. We use a pooling operator for a compact representation as follows:
\be
\bg(t) = \text{POOL}\left(\cE_{\theta_{\text{GNN}}}(\bx(t), \cE(t), \bW(t))\right), \label{eq:pool}
\ee
where \text{POOL$(\cdot$)} represents an average pooling function. 
\subsubsection{KAE} The KAE uses the input graph embedding, which is presumably non-linear due to the non-linearity inherent in the original dynamics and the graph encoder. With processable input of the graph embedding as the state vector of the dynamical system, we linearize using \eqref{EQ:enc} and the future embedding predictions is found using the Koopman matrix, as seen in \eqref{EQ:pred} and \eqref{EQ:dec}. 

\subsubsection{Graph Decoder } A graph decoder aims at reconstructing from the graph embedding state to the original graph state. Using the input as the graph embedding at any time step, we can reconstruct the node features as 
\be
\hat{\bx}(t+1) = \cD_{\phi_{\text{GNN}}}(\bg(t+1)), 
\ee
where using the reconstructed node features, the adjacency matrix can be derived based on a predefined rule for edge formation. If this rule is not available, methods such as those in \cite{kipf2016variational} can be employed to reconstruct the edges. In most cases, including this work, we assume that the method for edge formation in the graphs is predefined, which is defined on the graph signals. The proposed architecture of the GKAE and the operations of each of its components is depicted in Fig.~\ref{fig:Proposed}, in which the GKAE's forward processing is given as:
\begin{align}
\notag G(t) &\xrightarrow{\text{POOL}(\mathcal{E}_{\theta_{\text{GNN}}}(\cdot))}  \mathbf{g}(t) \xrightarrow{\mathcal{E}_{\theta_{\text{KAE}}}(\cdot)} \mathbf{h}(t) \\
 &\xrightarrow{\mathbf{K}} \mathbf{h}(t+1) \xrightarrow{\mathcal{D}_{\phi_{\text{KAE}}}(\cdot)} \mathbf{g}(t+1) \xrightarrow{\mathcal{D}_{\phi_{\text{GNN}}}(\cdot)} \bx(t+1).
\end{align}

\textbf{Training and Loss Function:} When training the GKAE model, we approximate a linear representation of a non-linear dynamical system using the Koopman matrix. Since the input is a spatio-temporal graph, we first convert it into a compact  lower-dimensional representation, which is the graph embedding. The non-linear dynamical system is modeled through the graph embedding, which captures information from multiple nodes in a lower-dimensional space, as illustrated in Fig.~\ref{fig:Proposed}. 
Thus, during training, our objective is to develop graph embeddings that retain minimal loss of information. These embeddings are subsequently transformed into the Koopman invariant subspace by the Koopman encoder to obtain a linear model up to $L$ time steps. The Koopman decoder then maps the Koopman invariant space back into the graph embedding space. Following this, we train the graph decoder to reconstruct the graph signals, enabling recovery of the original spatio-temporal graph.

To minimize the information loss during the transformation from the spatio-temporal graph to the graph embedding and back, we incorporate a reconstruction loss, which is defined as
\begin{align}
    L_{\text{GNN}} &= \sum_{t = 0}^{T-1} ||\mathbf{x}(t) - \hat{\mathbf{x}}(t)||^2, \label{eq:rec1}
\end{align}
Using \eqref{eq:rec1} and \eqref{eq:KAE}, the aggregated loss function for the GKAE is given by:
\begin{equation}
    \mathcal{L}_{\text{GKAE}}(\theta, \bK, \phi, L) = L_{\text{GNN}} + {\cL}_{\text{KAE}}(\theta_{\text{}}, \phi_{\text{}}, \bK, L), \label{eq:lossall}
\end{equation}
We use a sufficiently large value for $L$ during training to ensure effective linear modeling of the graph embedding.

\textbf{Linear Predictions using GKAE: } The GKAE can be used to predict future time-varying graph signals in a linear subspace. This is quite straightforward due to its simple architecture. 
 
We use $P$ to denote the total number of steps, we aim to predict linearly. The value of $P$ is determined based on the specific application requirements and is distinct from $L$, which represents the number of steps during the training to ensure linear modeling of the graph embeddings. The graph embedding for the spatio-temporal graph at the first time step is given as 
\be
\bg(0) = \cE_{\theta_\text{GNN}}(\bx(t), \cE(t), \bW(t)).
\ee
It is assumed that reconstructing the graph signals directly using the graph decoder should result in minimal reconstruction error, thereby ensuring that the learned embeddings effectively preserve the spatial characteristics of the original graph. The $p-$th time step prediction of the graph signals is now given using:
\be
\hat{\mathbf{x}}(p) = \mathcal{D}_{\phi_{\text{GNN}}}\biggl(\mathcal{D}_{\phi_{\text{KAE}}}\bigl(\mathbf{K}^{p} (\mathcal{E}_{\theta_{\text{KAE}}}\mathbf{g}(0))\bigr)\biggr), \quad p = 1, 2, \cdots, P.  \label{eq:pred}
\ee


\section{GKAE Application 2: Time-Varying Graph Reconstruction} \label{sec:recon}
In this section, we use the GKAE to address data loss and reconstruct missing node information in time-varying graph signals.

\subsection{GKAE-based Approach}
The GKAE can be employed on limited historical data of $\tau$ time steps, which is fully observable 
\be
\mathbb{G} = [G(0), \cdots, G(\tau-1)],
\ee
where the objective is approximating a global linear model during training. With successful linear modeling, the future graph embeddings for unseen $p-$step predictions can be obtained as:
\be
\bg(\tau + p) = \cD_{\phi_{\text{KAE}}}(\bK^{p} (\cE_{\theta_{\text{KAE}}}\bg(\tau))), \ p = 1, \cdots, T - \tau - 1 \label{eq:ge}.
\ee
Beyond $\tau$, the graph signals are sampled and only partially observable. Using GKAE, we can predict the graph embedding, as seen in \eqref{eq:ge}, which is a representation of the fully observable graph signals in a latent space. The objective is to train a secondary autoencoder that maps the partially observable graph signals for each time step $t$ to these predicted embeddings. The encoder of this secondary autoencoder takes the partially observable graph as input, while the decoder reconstructs the original graph signal. By aligning the output of the secondary autoencoder with the GKAE-predicted embeddings, the framework enables the reconstruction of the missing values in the original graph signals. This approach ensures the latent representation learned from the partially observable data aligns with the latent representation of the fully observable graph signals.
\subsubsection{Extending to Sampled Graph Input}
We propose a latent consistency autoencoder (LC-autoencoder) to process sampled spatio-temporal graphs and reconstruct the complete graph signals by aligning them with predicted graph embeddings.

The LC-autoencoder consists of an encoder and decoder, which are denoted using $\cE_{\theta_{\text{LC}}}$ and $\cD_{\phi_{\text{LC}}}$, respectively, and are trained following the training of the GKAE. The encoder takes as input the sampled graph signals $\bY$ and map it to the predicted graph embedding. At time $t$, the masked input signals and corresponding adjacency matrix are defined as 
\begin{align}
\by'(t) &= \bJ(:, t) \odot \bx(t) \in \mathbb{R}^N\\
\bW'(t) &= \bW(t) \odot (\bJ(:, t)\bJ(:, t)^\top) \in \mathbb{R}^{N \times N}
\end{align}
The encoder function of the LC-autoencoder takes the sampled input signals and the adjacency matrix and outputs a graph embedding, which is denoted using $\bar{\bg}(t)$. The graph embedding is the input to the decoder of the LC-autoencoder, which reconstructs the complete graph signal using the decoder function. To achieve this, we enforce consistency in the encoder and decoder functions with the trained GKAE model. Since the input to the LC-autoencoder encoder differs from that of the GKAE model, consistency is ensured in the graph embedding rather than the parameters of the encoder. Once the same graph embedding is achieved with the sampled spatio-temporal graph, consistency between the parameters of the GKAE decoder and the LC-autoencoder decoder allows reconstruction of the complete graph signal. The flow of the LC-autoencoder is defined as:
\begin{align}
\bar{\bg}(t) &= \cE_{\theta_\text{LC}}(\by'(t), \bW'(t)), \\
\bar{\bx}(t) &= \cD_{\phi_\text{LC}}(\bar{\bg}(t)), \label{eq:recon}
\end{align}
\subsubsection{Training}


Our overall training approach combines inductive pre-training with a transductive reconstruction phase. More specifically, it consists of two steps:
\begin{itemize}
    \item \textbf{GKAE Pre-training:} The GKAE is trained on the fully observed training set (i.e., time steps $0$ to $\tau-1$) to model system dynamics and predict future latent representations. The GKAE is trained using the loss function as in \eqref{eq:lossall}. 
    
    \item \textbf{Transductive Reconstruction:} For this task, the LC-autoencoder is trained on the masked version of the test set. The pre-trained GKAE provides predicted target embeddings for this period, as seen in \eqref{eq:ge}. The LC-autoencoder learns to map the incomplete input to these targets and reconstruct the corresponding graph signals. This process leverages information (the GKAE targets) derived from the unmasked test set itself, enabling accurate inpainting of missing values without generalizing to new, unseen masked datasets.
\end{itemize}
The parameters of these functions are optimized using the following loss functions:
\begin{align}
    L_{\text{latent}} &= \sum_{t = 0}^{T-1} \|\bar{\bg}(t) - \bg(t)\|^2, \\
    L_{\text{cosine}} &= \sum_{t = 0}^{T-1} \biggl| 1 - \frac{\langle \bar{\bg}(t), \bg(t) \rangle}{\|\bar{\bg}(t)\| \|\bg(t)\|} \biggl|, \\ 
    L_{\text{par}} &= \sum_{t = \tau}^{T-1} \|\cD_{\phi_{\text{LC}}}(\bar{\bg}(t)) - \cD_{\phi_{\text{GNN}}}(\bg(t))\|^2,
\end{align}
where $L_{\text{latent}}$ and $L_{\text{cosine}}$ enforce \emph{latent variable consistency}, while the term $L_{\text{par}}$ ensures \emph{decoder output consistency}. Therefore, even if some input data is missing due to masking or sampling limitations, as long as the same graph embedding can be generated, the sequences of unseen data can still be accurately reconstructed.

The final loss function for training the LC-autoencoder is given as:
\begin{equation}
\cL_{\text{LC}} = \beta_1 (L_{\text{latent}} + L_{\text{cosine}}) + \beta_2 L_{\text{par}},
\end{equation}
where $\beta_1$ and $\beta_2$ are hyperparameters controlling the relative importance of each loss component.

\begin{figure}
    \centering
    \includegraphics[height = 50mm, width=0.5\textwidth]{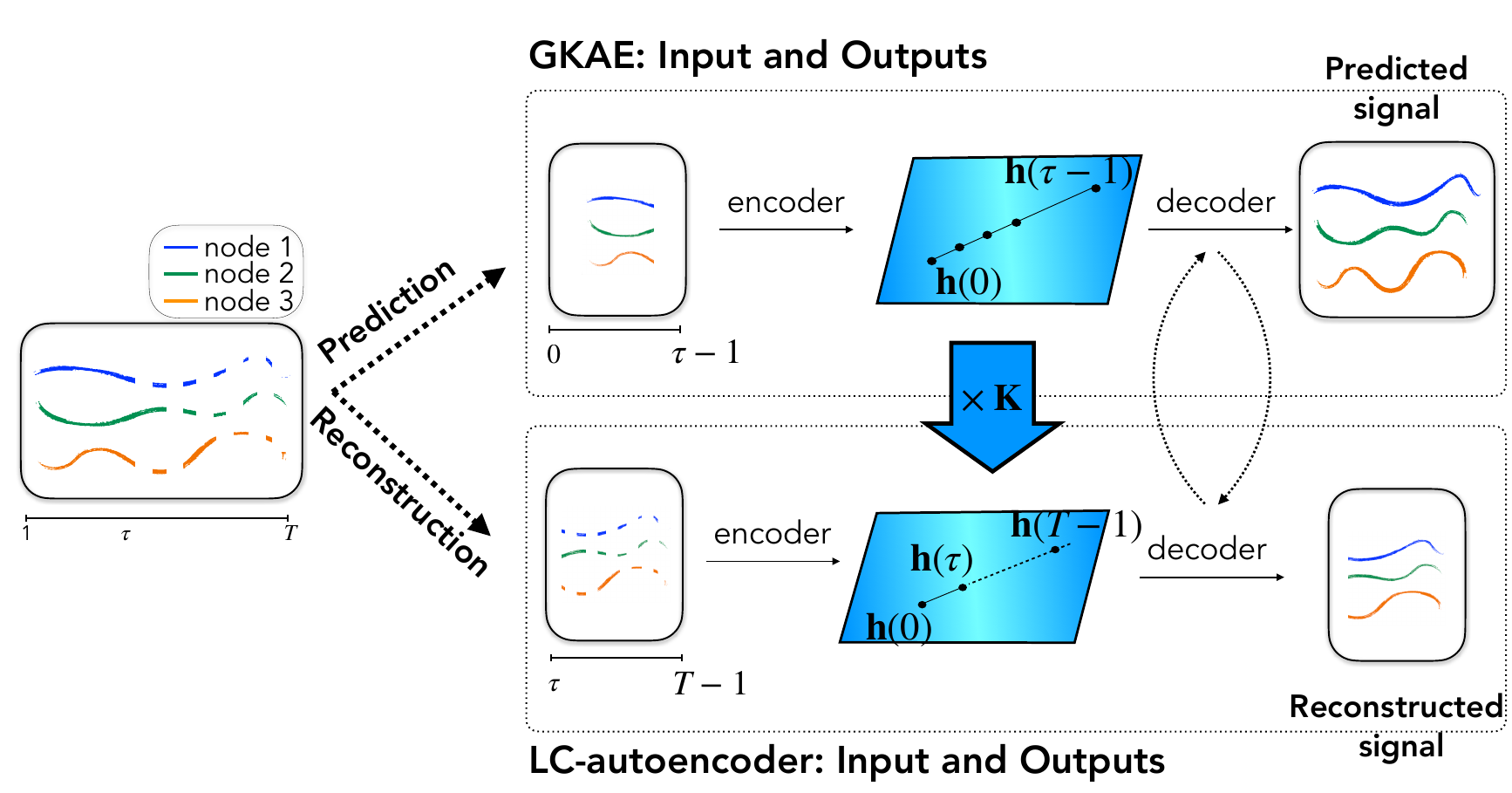}
\caption{Application of the GKAE for prediction and reconstruction using the LC-autoencoder, which processes sampled graph data while maintaining latent space consistency to reconstruct the original graph signals.}
    \label{fig:}
\end{figure}

\section{Simulation Setup and Results} \label{sec:SSR}
\subsection{Baseline Methods}
\paragraph{Reconstruction}
State-of-the-art solutions for reconstructing graph signals include GSP techniques. Other than this, we also include \emph{vanilla} GCN \cite{kipf2016semi} and nearest neighbor interpolation (NNI) \cite{sibson1981brief} as baselines. We consider the following baselines for comparison:
\begin{itemize}
    \item \textbf{Temporal graph smoothness (TGS)} \cite{Qiu17}: This approach leverages temporal smoothness constraints to reconstruct graph signals effectively. This method was trained using a gradient projection algorithm with backtracking line search, for up to 500 outer iterations or until convergence of the gradient norm. Each outer iteration adaptively selects a step size through at most 40 internal backtracking steps. 
    \item \textbf{Temporal graph Sobolev smoothness (TGSS)} \cite{giraldo2022reconstruction}: This approach extends TGS by incorporating Sobolev smoothness, capturing both temporal smoothness and higher-order variations for enhanced reconstruction. We use the same training settings as the TGS method. 
    \item \textbf{D\_X\_L$_2$ / S\_X\_L$_2$(\cite{yamagata2022robust})}: This approach extends graph signal reconstruction over Type-2 and Type-3 time-varying graphs. Both the approaches D\_X\_L$_2$ (dynamic graph structure) and S\_X\_L$_2$ (static graph structure) are trained for 2000 iterations. 
\end{itemize}
Other approaches include:
\begin{itemize}
    \item NNI \cite{sibson1981brief}. 
    \item Vanilla GCN \cite{kipf2016semi}: We train a encoder-decoder architecture with 4 GCN layers. The method is trained for 500 epochs.
\end{itemize}

\begin{table}[h!]
    \centering
    \caption{Simulation parameters}
    \begin{tabular}{|c|c|}
        \hline
        \hline
        \multicolumn{2}{|c|}{\textbf{Model Architecture}} \\
        \hline
        \hline
        Size of Koopman matrix $M$ & 8 \\
        Size of graph embedding $b$ & 8 \\
        Epochs & 200 \\
        Hyperparameters  $\beta_1, \beta_2$ & 1, $10^{-2}$ \\
        Linearity length $L$ & 50 \\
        Prediction length $P$ & 200\\
        Masking ratio & 50\% \\
        Length of observable data $\tau$ & 300 \\
        Total length of data $T$ & 500 \\
        Optimizer & Adam \\
        Learning rate & $10^{-2}$ \\
        \hline
        \hline
        \multicolumn{2}{|c|}{\textbf{Type-1 Datasets}} \\
        \hline
        \hline        
        Number of nodes $N$ & 20. \\
        Number of neighbors $k$ & 4 \\
        \hline
        \hline        
        \multicolumn{2}{|c|}{\textbf{Type 3 Datasets}} \\
        \hline
        \hline
        Number of nodes $N$ & 20. \\
        Communication radius of UAVs $r$ &0.5 \\
        Transmit power & 10 dBm \\
        Path loss & 2 \\
        \hline
        \hline
    \end{tabular}
    \label{tab:params}
\end{table}

\paragraph{Prediction}
For graph signal prediction tasks, we consider two sets of baselines. \emph{Koopman-based} state-of-the-art methods primarily emphasize temporal dynamics but often overlook the spatial structure of graphs. We use the following methods as baselines:
\begin{itemize}
    \item \textbf{KAE} \cite{lusch2018deep}: Focuses on learning a Koopman-invariant subspace for linear modeling of temporal sequences. On average, the total number of parameters in this model is 3270. 
    \item \textbf{Consistent KAE (cKAE)} \cite{azencot2020forecasting}: Extends KAE by enforcing consistency in temporal evolution for better predictive performance by adding a backward prediction matrix. On average, the total number of parameters is slightly higher than the KAE. 
\end{itemize}
These methods concatenate the input states of nodes to predict their future states; however, their disregard for the spatial structure of the graph significantly limits their capacity to model spatio-temporal dependencies effectively.
While \emph{GNN-based} approaches include:
\begin{itemize}
    \item \textbf{Seq-2-Seq GNN with LSTM/GRU layers:} A graph-based encoder-decoder model that applies a GCN for spatial encoding and an LSTM or GRU for temporal modeling. Trained with approximately 18K–24K parameters.
    
    \item \textbf{STGCN} \cite{yu2017spatio}: A spatio-temporal model using two ST-Conv blocks \cite{yu2017spatio}, each with spatial graph convolutions and temporal convolutions. Trained with approximately 8K parameters.
    
    \item \textbf{DCRNN} \cite{li2017diffusion}: A recurrent architecture using one DCRNN cell in both the encoder and decoder, modeling spatial diffusion and temporal dynamics. Trained with approximately 8K parameters.
    
    \item \textbf{CITRUS/CGCN} \cite{einizade2024continuous}: We use a feed-forward model with a linear input layer, four residual blocks, and an output layer. Each block uses spectral graph convolution with learnable time parameters and a small MLP. Trained with approximately 5K parameters.
\end{itemize}
Unless specified otherwise, each baseline approach is trained for 500 iterations/epochs.
\subsection{Datasets}
We use several datasets for the empirical evaluation of our method against baseline approaches for both reconstruction and prediction tasks.
\begin{itemize}
\item \textbf{Type-1 time-varying graph:}
    \begin{itemize}
        \item \textbf{NOAA SST} \cite{noaa_sst}: Global sea surface temperature (SST) measurements sampled weekly. The graph is static, constructed using the $k$-nearest neighbors (KNN) based on spatial proximity. For prediction, 300 weeks are used for training and 200 for testing. All experiments are conducted using 20 randomly selected nodes.

        \item \textbf{METR-LA} \cite{li2017diffusion}: Traffic speed data from loop detectors in Los Angeles, sampled every 5 minutes. Nodes represent sensors, and edges are defined by physical road connectivity. For prediction, 10 days are used for training and 4 days for testing.

        \item \textbf{PEMS-BAY} \cite{li2017diffusion}: Similar to METR-LA, with traffic speed readings from the Bay Area. The same prediction split is used. 

        \item \textbf{Beijing Air Quality} \cite{beijing_multi-site_air_quality_501}: Meteorological and air quality data collected hourly from monitoring stations in Beijing. Nodes represent stations, with edges based on spatial proximity. We use atmospheric pressure (hPa) as the graph signal. For both prediction, 300 hours are used for training and 200 for evaluation.
    \end{itemize}

    \item \textbf{Type-3 time-varying graph:} 
    \textbf{UAV} \cite{shao2023path}: Simulated data of mobile UAVs, where the graph structure evolves over time as UAVs change neighbors based on mobility within a radius $r$. The graph signal is the average signal-to-noise ratio (SNR), sampled at 0.1-second intervals. For both prediction, 30 seconds are used for training and 20 seconds for evaluation.
\end{itemize}

Note that since type-3 time-varying graph is a subset of type-2 time-varying graph, we do not consider a dataset of type-2 time-varying graph. We present results for both prediction and reconstruction tasks across the various datasets considered.
\subsection{Performance Metrics}
To evaluate the performance of the methods, we use the following metrics:

\begin{enumerate}
    \item \textbf{Prediction Error:} The root mean-squared error (RMSE) and mean absolute error (MAE) for predictions are defined as:
    \begin{align}
    \text{RMSE}_{\text{pred}} &= \sqrt{\frac{1}{P} \sum_{p = 1}^P \|\mathbf{x}(p) - \hat{\mathbf{x}}(p)\|^2}, \\
    \text{MAE}_{\text{pred}}  &= \frac{1}{P} \sum_{p = 1}^P \|\mathbf{x}(p) - \hat{\mathbf{x}}(p)\|_1,
    \end{align}
    where $\hat{\mathbf{x}}(p)$ is the predicted graph signal at time step $p$, and $P$ is the number of predictions.
    \item \textbf{Reconstruction Error:} The reconstruction loss, calculated over $T - \tau - 1$ steps on the masked input, is defined as:
    \be
    \epsilon_{\text{recon}} = \frac{1}{T - \tau - 1} \sum_{t = \tau}^{T - 1} \sum_{n:[\bJ]_{n, t}}\|\mathbf{x}_n(t) - \bar{\mathbf{x}}_n(t)\|^2,
    \ee
    where $\bar{\mathbf{x}}(t)$ is computed as described in \eqref{eq:recon}.
\end{enumerate}
\subsection{Architecture of the GKAE} 
We use the PyTorch geometric library \cite{fey2019fast} for simulating our GKAE architecture. 

\paragraph{GKAE} The graph encoder comprises two graph neural network layers: the first is a graph convolutional layer \cite{morris2019weisfeiler} that aggregates local neighborhood information using the graph structure and outputs a feature of dimension 8, and the second is a SAGEConv layer \cite{hamilton2017inductive}, which performs inductive node embedding by sampling and aggregating features from neighbors, also outputting features of dimension 8. Following the graph encoder, we apply a global mean pooling operator for a concise graph representation ($b = 8$). The Koopman autoencoder \cite{azencot2020forecasting} operates entirely within the latent space and comprises 12 fully connected layers—6 in the encoder and 6 in the decoder, each with 16 hidden nodes. The latent dynamics are modeled using a learnable Koopman operator represented as a fully connected linear transformation of dimension~$M \times M$ without an added bias term. The final reconstruction of the original graph signal is performed by the graph decoder which is composed of two fully connected layers applied to the Koopman-decoded latent space. We apply the Leaky ReLU activation function in both the graph encoder and decoder, while the Koopman autoencoder employs the hyperbolic tangent (tanh) activation in all its fully connected layers. The total number of parameters for training the GKAE model is approximately 4K parameters. 

\paragraph{LC-autoencoder} The LC-autoencoder is a secondary model specifically trained to handle the reconstruction of incomplete data. We use 4 layers, which is equally divided between the encoder and the decoder. The encoder processes the masked input with SAGECONV layers \cite{hamilton2017inductive}, which is followed by a global pooling operator to generate the same graph embedding of dimension $b$. We use the Leaky ReLu activation function in this architecture as well. The LC-autoencoder consists of approximately 3K parameters.
\begin{figure}[!h]
    \centering
    \includegraphics[width=0.8\columnwidth]{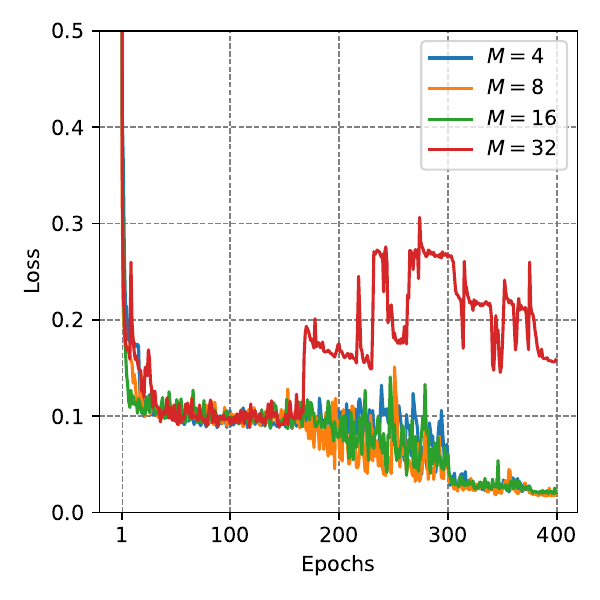}
    \caption{Loss convergence over varying dimensions for different Koopman matrix dimensions $M$.}
    \label{fig:varym}
\end{figure}

Unless specified otherwise, the parameters for defining our simulation environment is declared in Table.~\ref{tab:params}.

\subsection{Results}
\begin{table*}[t]
    \centering
    \caption{Prediction performance across different models and datasets. Each cell reports the performance metric averaged over 10 evaluation runs which is evaluated on the test set.}
    \label{tab:comparison_metrics_last_horizon}
    \resizebox{\textwidth}{!}{%
    \begin{tabular}{c|cc|cc|cc|cc|cc}
        \toprule
        \multirow{2}{*}{\textbf{Model}} 
        & \multicolumn{2}{c|}{\textbf{SNR (2s)}} 
        & \multicolumn{2}{c|}{\textbf{SST (6 months)}} 
        & \multicolumn{2}{c|}{\textbf{METR-LA (60m)}} 
        & \multicolumn{2}{c|}{\textbf{PEMS-BAY (60m)}}
        & \multicolumn{2}{c}{\textbf{pMRA Pressure (24h)}} \\
        
        \cmidrule(lr){2-3} \cmidrule(lr){4-5} \cmidrule(lr){6-7} 
        \cmidrule(lr){8-9} \cmidrule(lr){10-11}
        & \textbf{RMSE} & \textbf{MAE} 
        & \textbf{RMSE} & \textbf{MAE} 
        & \textbf{RMSE} & \textbf{MAE} 
        & \textbf{RMSE} & \textbf{MAE} 
        & \textbf{RMSE} & \textbf{MAE} \\
        \midrule

        KAE \cite{lusch2018deep} & 
        0.79 & 0.53 & 
        1.26 & 0.86 & 
        8.96 & 6.56 & 
        5.57 & 3.59 & 
        4.28 & 3.65  \\

        cKAE \cite{azencot2020forecasting} & 
        0.30 & 0.20 & 
        2.56 & 1.84 & 
        8.98 & 6.50 & 
        5.53 & 3.52 & 
        4.42 & 3.78 \\

        GCN-GRU & 
        0.23 & 0.19 & 
        2.71 & 1.91 & 
        9.30 & 7.05 & 
        5.37 & 3.51 & 
        4.54 & 4.03 \\

        GCN-LSTM & 
        0.24 & 0.18 & 
        1.52 & 1.20 & 
        9.30 & 6.80 & 
        5.44 & 3.40 & 
        4.01 & 3.60 \\

        STGCN \cite{yu2017spatio} & 
        0.51 & 0.37 & 
        3.34 & 2.37 & 
        8.71 & 5.94 & 
        5.08 & 3.13 & 
        4.89 & 4.08 \\

        cGCN \cite{einizade2024continuous} & 
        0.43 & 0.28 & 
        4.41 & 3.80 & 
        11.8 & 9.62 & 
        8.25 & 6.40 & 
        6.35 & 5.79 \\

        DCRNN \cite{li2017diffusion} & 
        0.30 & 0.23 & 
        1.95 & 1.36 & 
        8.42 & 5.81 & 
        5.63 & 3.60 & 
        5.82 & 5.04 \\
    \midrule \midrule
        GKAE (Ours) & 
        \textbf{0.11} & \textbf{0.06} & 
        \textbf{1.12} & \textbf{0.78} & 
        \textbf{8.31} & \textbf{5.78} & 
        \textbf{4.12} & \textbf{2.88} &
        \textbf{3.79} & \textbf{3.32} \\
        
        \bottomrule
        \bottomrule
    \end{tabular}%
    }
    \vspace{0.5em}
\end{table*}
\begin{figure*}[h!]
    \centering
    \begin{subfigure}[b]{0.3\textwidth}
        \centering
        \includegraphics[width=\textwidth]{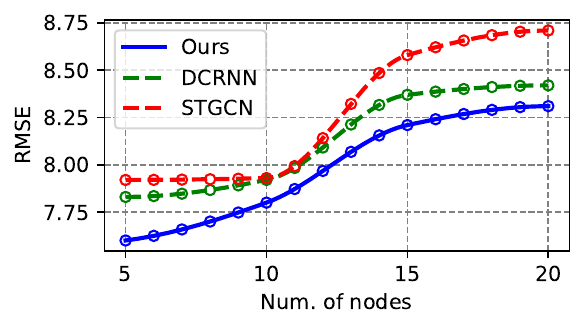}
        \caption{RMSE comparison on a Traffic dataset: METR-LA}
        \label{fig:rmse1}
    \end{subfigure}%
    \hfill
    \begin{subfigure}[b]{0.3\textwidth}
        \centering
        \includegraphics[width=\textwidth]{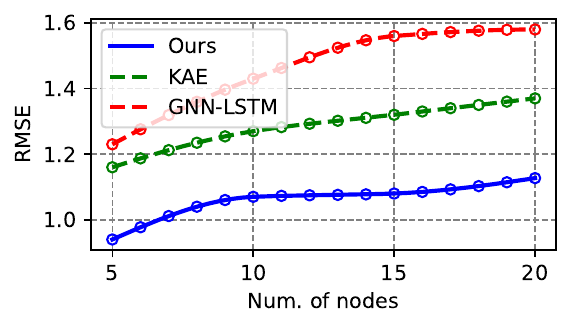}
        \caption{RMSE comparison on SST NOAA dataset}
        \label{fig:rmse2}
    \end{subfigure}%
    \hfill
    \begin{subfigure}[b]{0.3\textwidth}
        \centering
        \includegraphics[width=\textwidth]{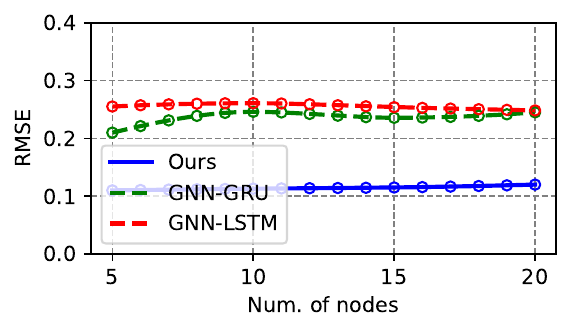}
        \caption{RMSE comparison on UAV-SNR dataset}
        \label{fig:rmse3}
    \end{subfigure}

    \begin{subfigure}[b]{0.3\textwidth}
        \centering
        \includegraphics[width=\textwidth]{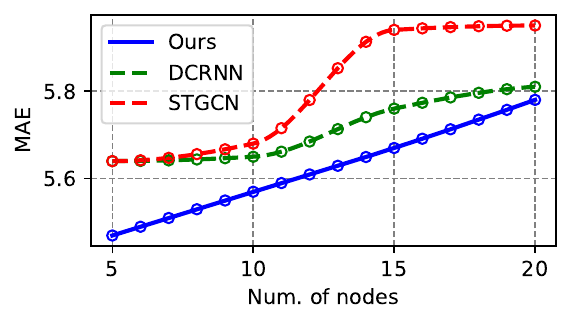}
        \caption{MAE comparison on METR-LA Traffic dataset}
        \label{fig:mae1}
    \end{subfigure}%
    \hfill
    \begin{subfigure}[b]{0.3\textwidth}
        \centering
        \includegraphics[width=\textwidth]{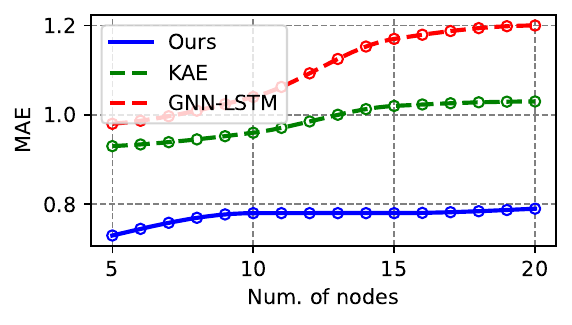}
        \caption{MAE comparison on SST NOAA dataset}
        \label{fig:mae2}
    \end{subfigure}%
    \hfill
    \begin{subfigure}[b]{0.3\textwidth}
        \centering
        \includegraphics[width=\textwidth]{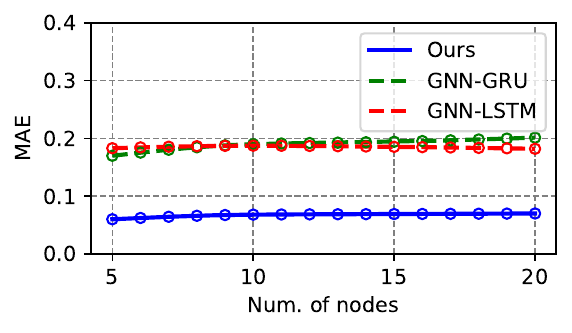}
        \caption{MAE comparison on UAV synthetic dataset}
        \label{fig:mae3}
    \end{subfigure}

    \caption{Performance comparisons over three dataset types over varying number of nodes $N$, compared for (top row) RMSE, (bottom row) MAE. The best three models have been compared.}
    \label{fig:perf_comparison}
\end{figure*}

\begin{figure*}[h]  
    \centering

    \begin{subfigure}[b]{0.30\textwidth}
        \centering
        \includegraphics[width=\textwidth]{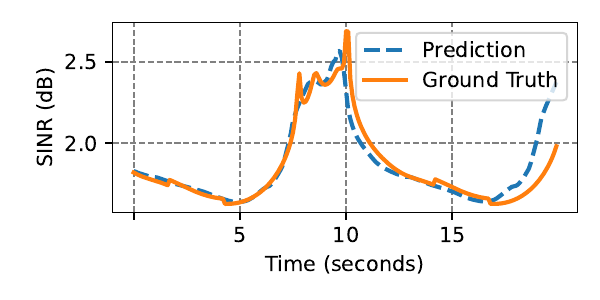}
        \caption{30-second horizon prediction on the UAV-SNR dataset.}
        \label{fig:pred1}
    \end{subfigure}
    \hfill
    \begin{subfigure}[b]{0.30\textwidth}
        \centering
        \includegraphics[width=\textwidth]{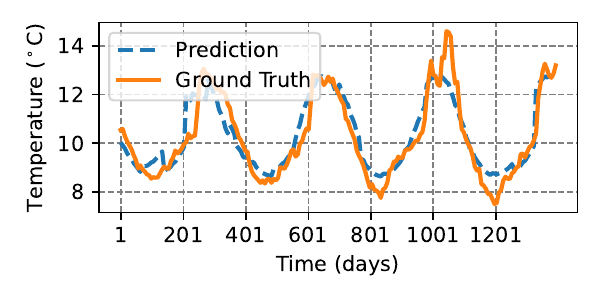}
        \caption{30-day horizon prediction of sea surface temperature (SST).}
        \label{fig:pred2}
    \end{subfigure}
    \hfill
    \begin{subfigure}[b]{0.30\textwidth}
        \centering
        \includegraphics[width=\textwidth]{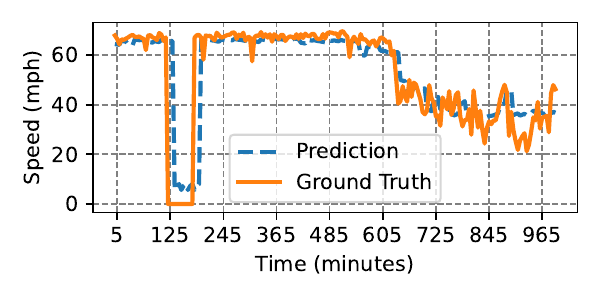}
        \caption{1-minute horizon speed prediction (METR-LA).}
        \label{fig:pred3}
    \end{subfigure}
    \hfill
    \begin{subfigure}[b]{0.30\textwidth}
        \centering
        \includegraphics[width=\textwidth]{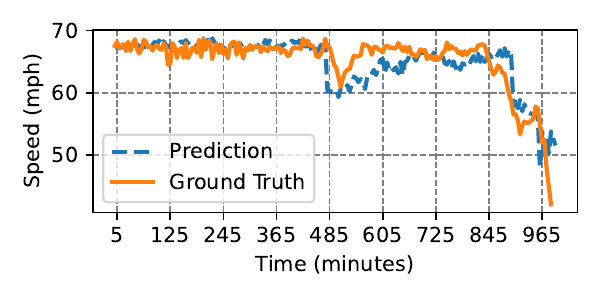}
        \caption{1-hour horizon speed prediction (PEMS-BAY).}
        \label{fig:pred4}
    \end{subfigure}
    \begin{subfigure}[b]{0.30\textwidth}
        \centering
        \includegraphics[width=\textwidth]{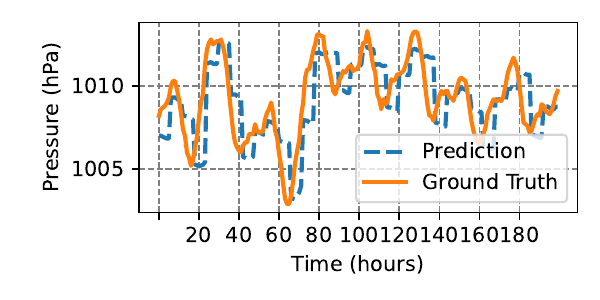}
        \caption{6-hour horizon pressure prediction in the Beijing dataset.}
        \label{fig:pred5}
    \end{subfigure}
    \caption{200-step prediction comparison across datasets with varying forecasting horizons, showing ground truth and predicted trajectories.}
    \label{fig:prediction_comparisons}
\end{figure*}

\subsubsection{Prediction via Graph Koopman Autoencoder}
In this section, we present the results for predicting $P$ steps over various datasets, which include synthetic dataset (UAV-SNR), traffic prediction dataset (METR-LA \& PEMS-BAY) and environmental datasets (NOAA SST \& Beijing Air Quality).

\textbf{Best Koopman dimension:} Firstly, in Fig.~\ref{fig:varym}, we compare the loss during training over varying dimension of the $\bK$ matrix. The dimension of the $\bK$ matrix is critical as it determines the dimensionality of the Koopman invariant subspace, which directly influences the model's ability to capture and predict the underlying dynamics. A matrix that is too small may fail to capture the necessary complexity of the system, leading to inaccurate predictions, while an excessively large matrix may introduce redundancy and increase computational overhead. A high dimensional $\bK$ matrix may also cause instability during training, particularly with long trajectory lengths due to recursive multiplication, as seen in the case with $M = 32$. 

In Table~\ref{tab:comparison_metrics_last_horizon}, we use a Koopman dimension of $M = 8$ for training our models and compare them against the aforementioned baselines for long-horizon predictions. 

On the UAV-SNR dataset, we forecast 2 seconds into the future, where our model significantly outperforms all baselines. Specifically, it achieves a 52.2\% reduction in RMSE and a 66.7\% reduction in MAE compared to GCN-LSTM.

On the SST dataset, our method improves long-range (6-month) predictions, achieving an 11.1\% reduction in RMSE and a 9.3\% reduction in MAE. While GCN-LSTM also shows competitive performance, our model outperforms it by 26.3\% in RMSE and 35\% in MAE.

For the more challenging traffic forecasting tasks, GKAE also demonstrates notable improvements. On the PEMS-BAY dataset, it reduces RMSE by 18.9\% and MAE by 8.0\% compared to the next-best model, STGCN. Although the performance margin is narrower on the METR-LA dataset, our model still surpasses all baselines.

Finally, on the pressure dataset, our approach outperforms GCN-LSTM by 5.5\% in RMSE and 7.8\% in MAE.
We also evaluate the RMSE and MAE over a varying number of nodes in the data for the three best-performing models on each type of dataset in Fig.~\ref{fig:perf_comparison}. Firstly, as depicted in Fig.~\ref{fig:rmse1} and Fig.~\ref{fig:mae1}, our model outperforms the DCRNN by 1.31-2.94\% in RMSE and 0.52-3.01\% in MAE, on average. The performance improvement is more pronounced in the SST and UAV-SNR datasets, where our model outperforms the next-best models by a significant margin. Specifically, on the SST dataset, it improves upon the GCN-LSTM baseline by 15.7-19.0\% in RMSE and 18.8-23.5\% in MAE, represented in Fig.~\ref{fig:rmse2} and Fig.~\ref{fig:mae2}, respectively. The advantage is even more substantial on the UAV-SNR dataset. It is shown in Fig.~\ref{fig:rmse3} and Fig.~\ref{fig:mae3} that our model outperforms the strongest baselines by 47.6-54.1\% in RMSE and 61.5-68.1\% in MAE.

In Fig.~\ref{fig:prediction_comparisons}, we compare the predicted trajectories against the ground truth over the test set for the specified horizon lengths, which shows that the GKAE can perform well for predicting on different types of datasets. 

\subsubsection{Reconstruction using LC-Autoencoder} 
For the reconstruction task, our objective is to accurately inpaint missing node information within time-varying graph signals. We employ the LC-autoencoder for this purpose, which reconstructs masked graph signals by aligning their latent space and decoder output with the representations provided by the pre-trained GKAE model. The detailed training and evaluation methodology for the LC-autoencoder, including its transductive learning framework and specific data utilization, is presented in Section~\ref{sec:recon}. Reconstruction performance is evaluated on its ability to reconstruct the masked graph signals in the test set. The following section presents a detailed comparison of reconstruction errors across multiple methods and datasets.
\begin{figure}[h]
    \centering
    \includegraphics[width=1\columnwidth]{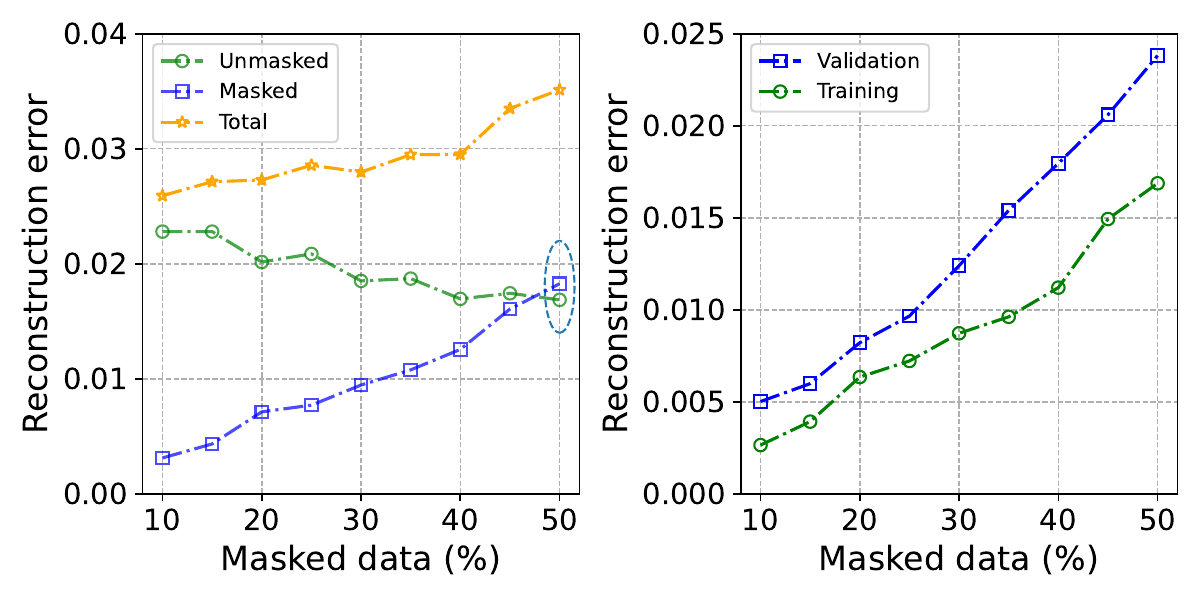}
    \caption{Evaluating the generalization ability of the GKAE for reconstruction by comparing the reconstruction of masked/unmasked nodes (left) and training/validation data (right).}
    \label{fig:re2}
\end{figure}

\begin{table*}[t]
    \centering
    \caption{Reconstruction performance ($\epsilon_\text{recon}$) across different masking rates and datasets.}
    \label{tab:reconstruction_metrics}
    \resizebox{\textwidth}{!}{%
    \begin{tabular}{c|ccccc|ccccc|ccccc|ccccc}
        \toprule
        \multirow{2}{*}{\textbf{Model}} 
        & \multicolumn{5}{c|}{\textbf{SNR (UAV)}} 
        & \multicolumn{5}{c|}{\textbf{SST}} 
        & \multicolumn{5}{c|}{\textbf{Beijing Pressure}} 
        & \multicolumn{5}{c}{\textbf{PEMS-BAY}} \\
        
        \cmidrule(lr){2-6} \cmidrule(lr){7-11} 
        \cmidrule(lr){12-16} \cmidrule(lr){17-21}
        & 10\% & 20\% & 30\% & 40\% & 50\%
        & 10\% & 20\% & 30\% & 40\% & 50\%
        & 10\% & 20\% & 30\% & 40\% & 50\%
        & 10\% & 20\% & 30\% & 40\% & 50\% \\
        \midrule

        TGS \cite{Qiu17} & 
        0.004 & 0.006 & 0.017 & 0.048 & 0.127 & 
        \textbf{0.052} & 0.477 & 0.633 & 1.047 & 2.057 &
        0.137 & 0.268 & 0.557 & 0.808 & 1.098 &
        54.62 & 162.63 & 284.75 & 430.46 & 585.78 \\

        TGSS \cite{giraldo2022reconstruction} & 
        0.003 & 0.010 & 0.023 & 0.046 & 0.099 & 
        0.123 & 0.572 & 0.773 & 1.191 & 1.431 &
        0.112 & 0.575 & 1.068 & 1.601 & 1.962 &
        65.96 & 114.82 & 151.61 & 180.46 & 214.85 \\

        D\_X\_L$_2$ \cite{yamagata2022robust} & 
        0.007 & 0.009 & 0.019 & 0.033 & 0.038 & 
        0.073 & 0.402 & 0.986 & 1.195 & 1.234 &
        0.074 & 0.169 & 0.231 & 0.326 & 0.377 &
        85.04 & 125.24 & 153.60 & 178.22 & 198.30 \\

        GCN \cite{kipf2016semi} & 
        0.088 & 0.208 & 0.312 & 0.456 & 0.590 &
        0.654 & 1.380 & 2.180 & 2.945 & 3.432 &
        0.703 & 1.557 & 2.635 & 3.814 & 4.479 &
        105.23 & 190.45 & 325.78 & 488.64 & 699.91 \\

        NNI \cite{sibson1981brief}  & 
        0.080 & 0.174 & 0.236 & 0.326 & 0.406 &
        0.944 & 2.012 & 3.341 & 4.167 & 4.985 &
        1.027 & 2.185 & 3.709 & 5.043 & 5.962 &
        123.61 & 207.78 & 341.09 & 512.83 & 742.50 \\

        S\_X\_L$_2$ \cite{yamagata2022robust}& 
        0.008 & 0.011 & 0.022 & 0.036 & 0.053 &
        -- & -- & -- & -- & -- &
        -- & -- & -- & -- & -- &
        -- & -- & -- & -- & -- \\

        \midrule\midrule
        GKAE (Ours) & 
        \textbf{0.002} & \textbf{0.003} & \textbf{0.012} & \textbf{0.030} & \textbf{0.034} &
        0.132 & \textbf{0.353} & \textbf{0.537} & \textbf{0.891} & \textbf{1.031} &
        \textbf{0.047} & \textbf{0.108} & \textbf{0.162} & \textbf{0.218} & \textbf{0.287} &
        \textbf{22.18} & \textbf{40.31} & \textbf{54.65} & \textbf{64.04} & \textbf{81.44} \\
        \bottomrule
    \end{tabular}%
    }
    \vspace{0.5em}
\end{table*}

\begin{figure*}[h]
    \centering
    \begin{subfigure}[b]{0.24\textwidth}
        \centering
        \includegraphics[width=\textwidth]{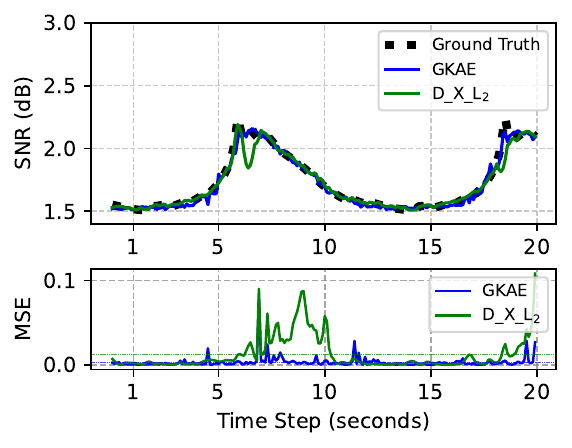}
        \caption{Reconstruction over the test set and the MSE over time on the UAV-SNR dataset for GKAE vs D\_X\_L2.}
        \label{fig:reSNR}
    \end{subfigure}%
    \hfill
    \begin{subfigure}[b]{0.24\textwidth}
        \centering
        \includegraphics[width=\textwidth]{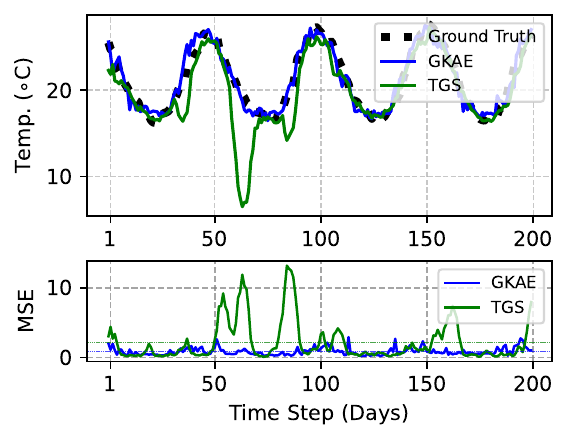}
        \caption{Reconstruction over the test set and the MSE over time on the SST dataset for GKAE vs TGS over 200 time steps.}
        \label{fig:reSST}
    \end{subfigure}%
    \hfill
    \begin{subfigure}[b]{0.24\textwidth}
        \centering
        \includegraphics[width=\textwidth]{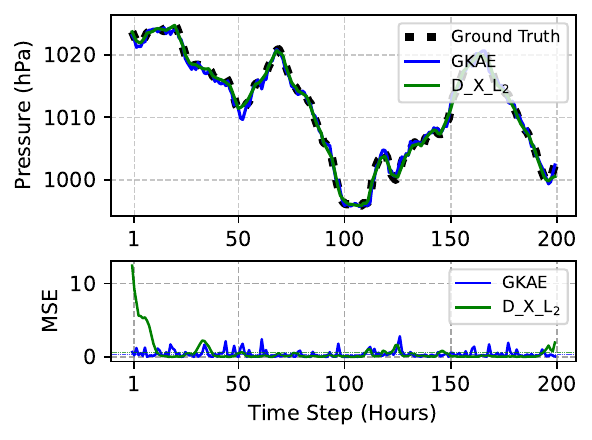}
        \caption{Reconstruction over the test set and the MSE over time on the pressure dataset for GKAE vs D\_X\_L2.}
        \label{fig:rePRES}
    \end{subfigure}%
    \hfill
    \begin{subfigure}[b]{0.24\textwidth}
        \centering
        \includegraphics[width=\textwidth]{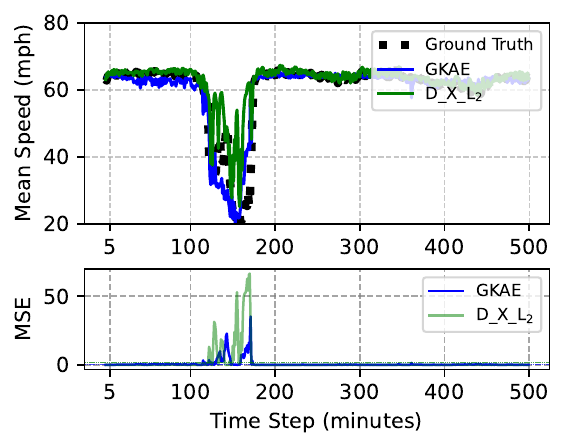}
        \caption{Reconstruction over the test set and the MSE over time on the PEMS-BAY dataset for GKAE vs D\_X\_L2.}
        \label{fig:rePEMS}
    \end{subfigure}
    \caption{Reconstruction comparison with MSE over time for the two best performing methods over varying datasets.}
    \label{fig:recon_comparison}
\end{figure*}

\begin{figure*}[h]
    \centering
    \begin{subfigure}[b]{0.24\textwidth}
        \centering
        \includegraphics[width=\textwidth]{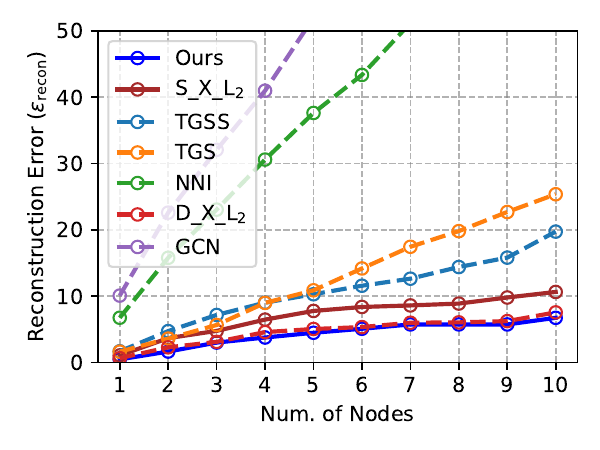}
        \caption{Reconstruction error for the various methods over varying number of nodes on the UAV-SNR dataset.}
        \label{fig:erSNR}
    \end{subfigure}%
    \hfill
    \begin{subfigure}[b]{0.24\textwidth}
        \centering
        \includegraphics[width=\textwidth]{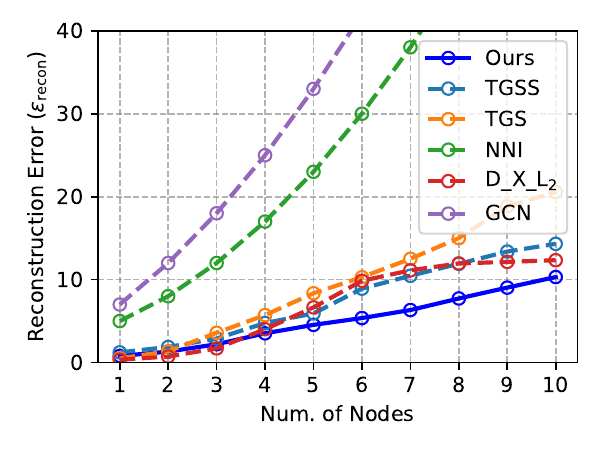}
        \caption{Reconstruction error for the various methods over varying number of nodes on the SST dataset.}
        \label{fig:erSST}
    \end{subfigure}%
    \hfill
    \begin{subfigure}[b]{0.24\textwidth}
        \centering
        \includegraphics[width=\textwidth]{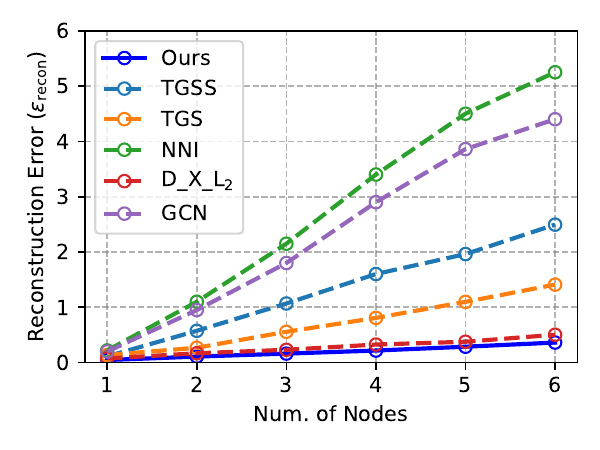}
        \caption{Reconstruction error for the various methods over varying number of nodes on the Beijing pressure dataset.}
        \label{fig:erPRES}
    \end{subfigure}%
    \hfill
    \begin{subfigure}[b]{0.24\textwidth}
        \centering
        \includegraphics[width=\textwidth]{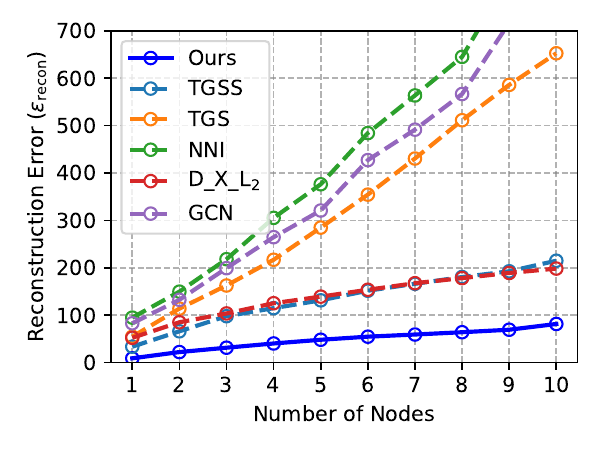}
        \caption{Reconstruction error for the various methods over varying number of nodes on the PEMS-BAY traffic dataset.}
        \label{fig:erPEMS}
    \end{subfigure}
    \caption{Reconstruction error over varying datasets for different methods.}
    \label{fig:errorcom}
\end{figure*}

\paragraph{Generalization}
In Fig.~\ref{fig:re2}, we evaluate the reconstruction capability of the LC-autoencoder. For this specific analysis of transductive inpainting, the figure presents results on a distinct subset of the data that serves to demonstrate generalization to unmasked or partially observed samples, labeled within the figure as the 'validation set.' This allows for an in-depth examination of the model's performance on both masked and unmasked nodes, as well as a comparison between its performance on the training data and this particular subset. 

The reconstruction error is compared for masked and unmasked nodes across different masking levels. While the error increases with higher proportions of masked data, it remains comparable when masking 50\% of the nodes, demonstrating the LC-autoencoder's ability to effectively reconstruct both visible and masked inputs. We also compare the reconstruction performance on the training and validation sets, noting that the validation reconstruction error is slightly higher than the training reconstruction error.

\paragraph{Comparison with Baselines on Varying Datasets} 
For the reconstruction task, we present results on multiple datasets across different masking rates, summarized in Table~\ref{tab:reconstruction_metrics}. It is worth noting that the $\epsilon_\text{recon}$ has been reported as the performance metric. It can be noted that our approach outperforms on almost every setting and dataset. In general, it can be seen that the TGSS \cite{giraldo2022reconstruction} and D\_X\_L$_{2}$ offer comparable performance in most of the scenarios. On average, our approach outperforms the TGSS by 57.4\%, while on the other hand, it surpasses the D\_X\_L$_2$ model by an average of 43.5\% across all applicable settings and datasets. he highest improvement was seen with a 64.9\% better performance than the TGSS model on the PEMS-BAY dataset at a 20\% masking rate, while the lowest was a 9.1\% improvement over the D\_X\_L$_2$ model on the SNR dataset at a 40\% masking rate. It is worth noting that the GCN and NNI do not offer good performance, while the TGS offers comparable performances when the masking rate is low. 

In Fig.~\ref{fig:recon_comparison}, we present reconstruction plots over the test set for the two best-performing methods on each dataset. For the UAV-SNR dataset, our approach is compared against D\_X\_L$_2$, which is designed to handle time-varying graph structures. Similarly, D\_X\_L$_2$ is used as a baseline for the Beijing pressure and PEMS-BAY traffic datasets, while TGS is used for comparison on the SST dataset.

On average, our model outperforms the next-best method by 12.4-56.4\% on the UAV-SNR dataset, 28.0-37.8\% on the SST dataset, 3.5-27.8\% on the Beijing pressure dataset, and 58.9-82.7\% on the PEMS-BAY dataset.

\paragraph{Completely Masked Nodes}
The previous results highlighted scenarios where masking is time-variant and sufficient unmasked data is available for each node. However, there may be instances during training where no information is available for certain nodes. Baseline methods, which rely on temporal smoothness to reconstruct missing values, require adequate unmasked input for effective training. In contrast, when a node is completely masked across all time instances, these methods struggle to accurately reconstruct the graph signals. Our approach, however, leverages spatial learning through the GNN to accurately predict the graph embedding. This capability allows our method to reconstruct the graph embedding even when a node remains entirely masked during training. 

\begin{figure}
    \centering
    \includegraphics[width=1\columnwidth]{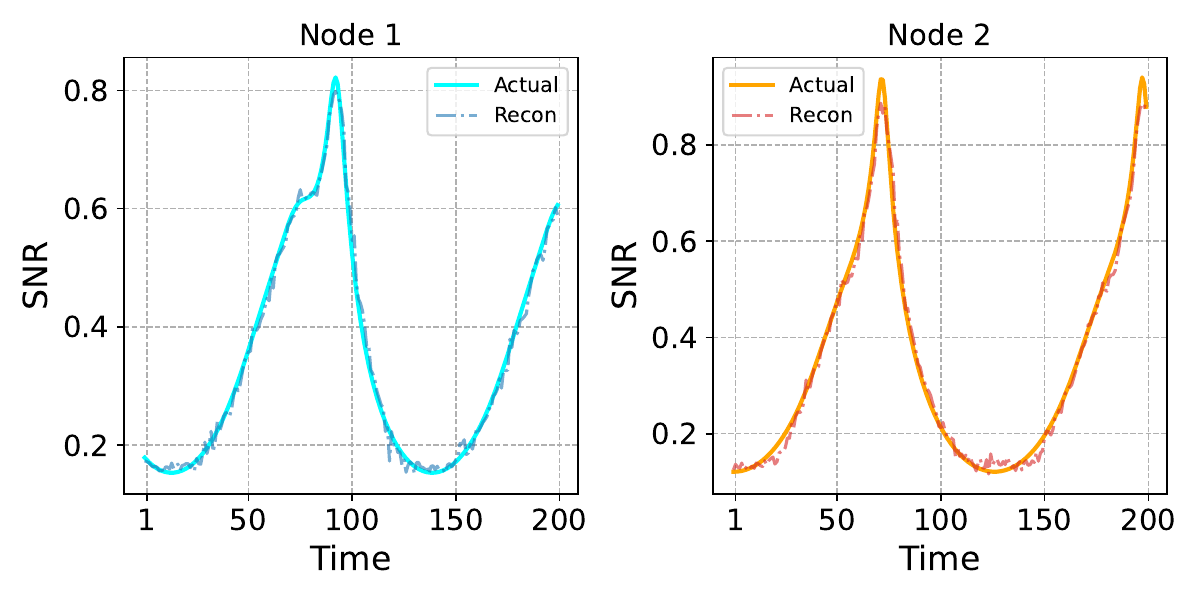}
    \caption{Reconstruction of graph signals when the graph signals for these nodes are completely masked during training.}
    \label{fig:singlerecon}
\end{figure}

In Fig.~\ref{fig:singlerecon}, we illustrate the reconstructed graph signals for two nodes that were completely masked during training. Despite the absence of any training data for these nodes, our method demonstrates the ability to reconstruct their signals with low error. This is achieved by leveraging the matching of latent variables through the graph embeddings, which capture the spatial relationships and dynamics of the graph. This result highlights the robustness of our approach in handling scenarios where no temporal information is available for certain nodes, effectively utilizing the graph structure and spatial learning to ensure accurate reconstructions.

\section{Conclusions} \label{sec:C}
In this work, we proposed a novel approach by leveraging GNNs and Koopman theory to address reconstruction and prediction tasks on time-varying graphs with dynamic structures. 
Unlike previous methods, which are typically restricted to time-varying graphs with static structures (e.g., where only node features vary over time while graph properties remain fixed, failing to capture dynamic relationships between nodes), our approach captured the evolving dynamics of node relationships through a flexible latent-space framework grounded in Koopman theory. This framework introduced a graph embedding technique applicable to any type of time-varying graph.
 
We evaluated the performance of our method against well-established baselines and demonstrated an average improvement of 57.4\% in reconstruction tasks and 18.1\% in long-term prediction tasks. Future work will focus on enhancing consistency and further improving long-term prediction accuracy.

\bibliographystyle{ieeetr}
\bibliography{Koopman}

\end{document}